%% file: main.tex
\pdfoutput=1
\documentclass[11pt]{article}
\usepackage[final]{acl}

\title{Nearest Neighbor Normalization Improves Multimodal Retrieval}

\author{
Neil Chowdhury\textsuperscript{1*}, Franklin Wang\textsuperscript{1*}, Sumedh Shenoy\textsuperscript{1*},\\{\bf Douwe Kiela\textsuperscript{2}},
{\bf Sarah Schwettmann{\textsuperscript{1}\textsuperscript{\textdagger}}}, {\bf Tristan Thrush\textsuperscript{2}\textsuperscript{\textdagger}} \\
\textsuperscript{1}Massachusetts Institute of Technology, 
\textsuperscript{2}Stanford University \\
\texttt{\{nchow,fxwang,sshenoy,schwett\}@mit.edu}, \texttt{\{dkiela,tthrush\}@stanford.edu} \\
\textsuperscript{*}Equal contribution
\textsuperscript{\textdagger}Equal advising
}

\input{packages}

\input{macros}

\begin{document}

\maketitle
\begin{abstract}
Multimodal models leverage large-scale pretraining to achieve strong but still imperfect performance on tasks such as image captioning, visual question answering, and cross-modal retrieval. In this paper, we present a simple and efficient method for correcting errors in trained contrastive image-text retrieval models with no additional training, called \textit{Nearest Neighbor Normalization} (\nnd). We show an improvement on retrieval metrics in both text retrieval and image retrieval for all of the contrastive models that we tested (CLIP, BLIP, ALBEF, SigLIP, BEiT) and for both of the datasets that we used (MS-COCO and Flickr30k). \nnd\ requires a reference database, but does not require any training on this database, and can even increase the retrieval accuracy of a model after finetuning.\footnote{Our code is publicly available at \url{https://github.com/multimodal-interpretability/nnn}}

\end{abstract}

\input{01_intro}
\input{02_method}
\input{03_experiments}

\input{04_conclusion}
\newpage
\input{045_ethics_limitations}

\bibliography{custom}
\clearpage
\appendix
\input{05_appendix}

\end{document}

%% file: packages.tex
\usepackage{times}
\usepackage{latexsym}

\usepackage[T1]{fontenc}

\usepackage[utf8]{inputenc}

\usepackage{microtype}

\usepackage{inconsolata}

\usepackage{amsmath}
\usepackage{amssymb}
\usepackage{mathtools}
\usepackage{amsthm}
\usepackage{booktabs}
\usepackage{multirow}
\usepackage{listings}
\usepackage{xcolor}
\usepackage{tcolorbox}
\tcbuselibrary{listings}
\usepackage{tcolorbox}
\usepackage{microtype}
\usepackage{graphicx}
\usepackage{subfigure}
\usepackage{float}
\usepackage{booktabs} 
\usepackage{amssymb}
\usepackage{pifont}
\usepackage{wrapfig}
\usepackage{caption}[=v1] 

%% file: macros.tex
\newcommand{\ie}[1]{\emph{i.e}}

\definecolor{boxgray}{rgb}{.94, .94, .94}

\lstdefinestyle{mystyle}{
  basicstyle=\ttfamily\scriptsize,
  breaklines=true,
  columns=fullflexible,
  breakindent=0pt,
}

\lstdefinestyle{mystylepython}{
    language=Python,
    basicstyle=\ttfamily\scriptsize,
    otherkeywords={self},             
    keywordstyle=\ttfamily\scriptsize\color{blue},
    emph={MyClass,__init__},          
    emphstyle=\ttfamily\scriptsize\color{blue},    
    stringstyle=\color{blue}, 
    commentstyle=\color{gray},   
    showstringspaces=false,            
    breaklines = true
    backgroundcolor=\color{lightgray}, 
}

\definecolor{codegreen}{rgb}{0,0.6,0}
\definecolor{codegray}{rgb}{0.5,0.5,0.5}
\definecolor{codepurple}{rgb}{0.58,0,0.82}
\definecolor{backcolour}{rgb}{0.95,0.95,0.92}

\newcommand{\code}[1]{\lstinline[language=Python]|#1|}

\newcommand{\nnd}{\code{NNN}}

\newcommand{\eg}{\emph{e.g.}}

%% file: 01_intro.tex
\section{Introduction}
\label{sec:intro}

Contrastive image and text models are a fundamental building block of large-scale text-to-image or image-to-text retrieval systems~\citep{radford2021learning, jia2021scaling, zhang2022contrastive}. These models utilize contrastive loss functions to learn joint text and image embeddings, aligning embeddings for matching text and image pairs while separating embeddings for non-matching pairs. However, contrastive embeddings optimize pretraining objectives such as InfoNCE \cite{radford2021learning} rather than downstream retrieval accuracy, so learned embeddings can be suboptimal for retrieval \cite{zhou2023testtime}. Many methods for improving contrastive models on downstream retrieval tasks require additional training to adapt models across domains or aggregate information from an external database \cite{zhou2022learning, singha2023ad, iscen2023retrievalenhanced}, and others are specialized for individual error categories, such as gender bias \cite{wang2021genderneutral, wang2022fairclip, berg2022prompt}.

\begin{figure}[t!]
    \centering
    \includegraphics[width=.82\linewidth]{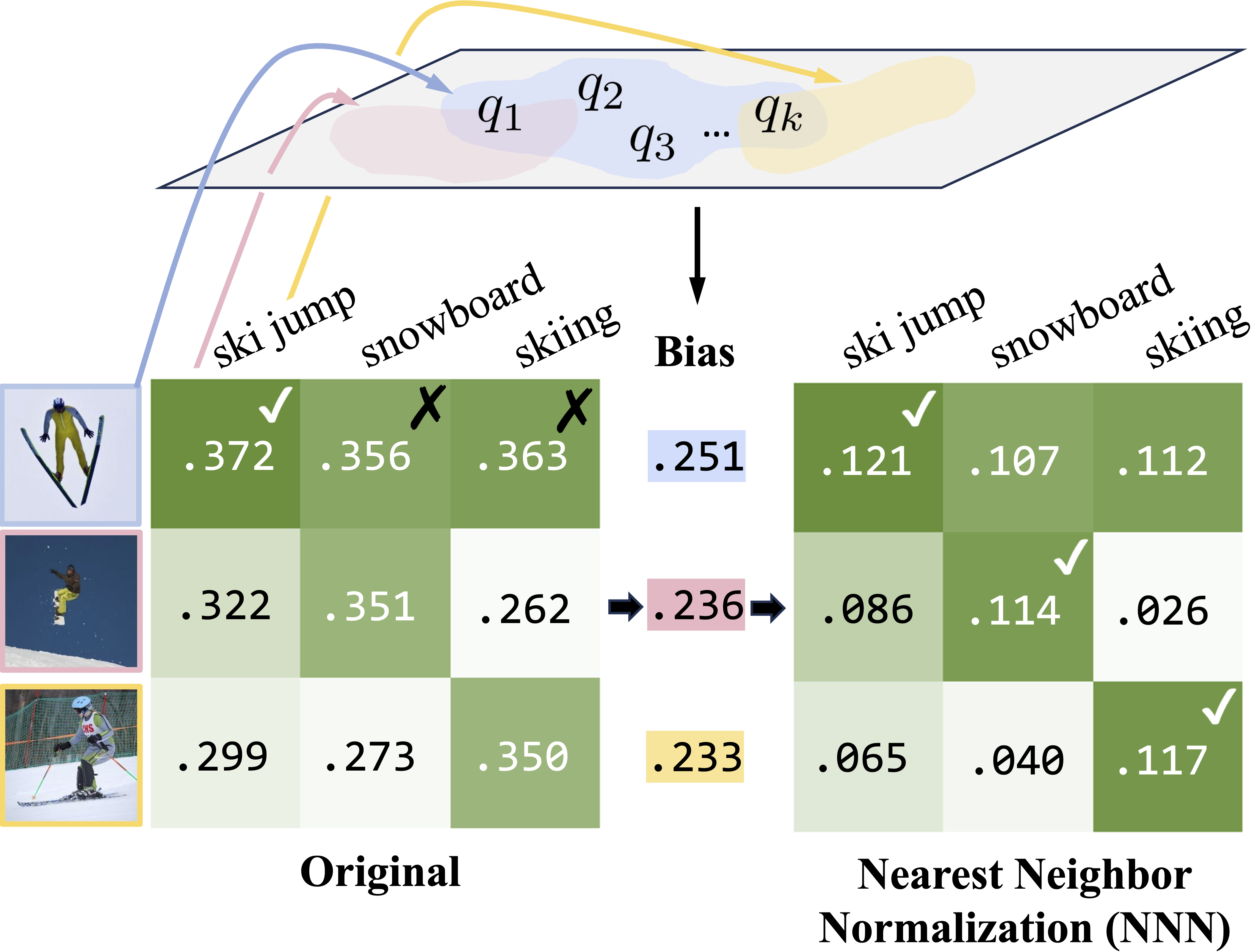}
    \vspace{-3mm}
    \caption{\textbf{Method overview.} \nnd\ applies an additive correction at inference time, using bias scores estimated from a reference database of queries.
    }
    \vspace{-7mm}
    \label{fig:teaser}
\end{figure}

Recent training-free methods suggest that accuracy can be improved without fine-tuning, which is useful for limited-compute environments and critical for black-box embedding models. Such methods typically use a reference database of query and retrieval embeddings to adapt the pretrained model to the downstream retrieval task. For instance, QBNorm and DBNorm normalize scores for each retrieval candidate by computing a softmax over the entire reference database \cite{bogolin2022cross, wang-etal-2023-balance}. These approaches mitigate the \textit{hubness} problem, where certain retrieval candidates (``hubs'') emerge as nearest neighbors for many queries in high-dimensional embedding spaces, leading to incorrect matches \citep{radovanovic2010hubs}. These methods tend to be computationally impractical, requiring match score calculations for every item in the database and thus scaling linearly with the size of the reference database. Distribution normalization (DN) reduces complexity to constant time by using a first-order approximation of softmax normalization \cite{zhou2023testtime}: text and image embeddings are normalized by subtracting the mean reference embedding. While DN is much faster than QBNorm and DBNorm, this practicality comes at the cost of reduced retrieval accuracy. \textbf{Can sublinear runtime be achieved without sacrificing accuracy?}

In this paper, we introduce Nearest Neighbor Normalization (\nnd), a novel training-free method for contrastive retrieval (Figure~\ref{fig:teaser}). Like DN, it adds minimal inference overhead with sublinear time complexity relative to the reference database size---but it also outperforms both QBNorm and DBNorm on retrieval. The key idea is that \nnd\ corrects for the effects of embeddings that are assigned disproportionately high or low retrieval scores, by normalizing per-candidate scores using \textit{only the $k$ closest query embeddings} from a reference dataset. 
For example, \nnd\ reduces scores for the image of the surfer in Figure~\ref{fig:matching} (a hub that incorrectly matches a large number of query captions), improving overall accuracy. 
Section \ref{sec:method} provides more details on our approach, and Section \ref{sec:experiments} empirically validates the effect of \nnd\ for a range of models and datasets.

Overall, we contribute a new and conceptually simple approach for improving contrastive retrieval with little compute overhead. In addition to improving retrieval scores consistently for every model and dataset that we tested, \nnd\ can reduce harmful biases such as gender bias. 

%% file: 02_method.tex
\section{Nearest Neighbor Normalization}
\label{sec:method}

\begin{figure}[t!]
    \centering
    \includegraphics[width=\linewidth]{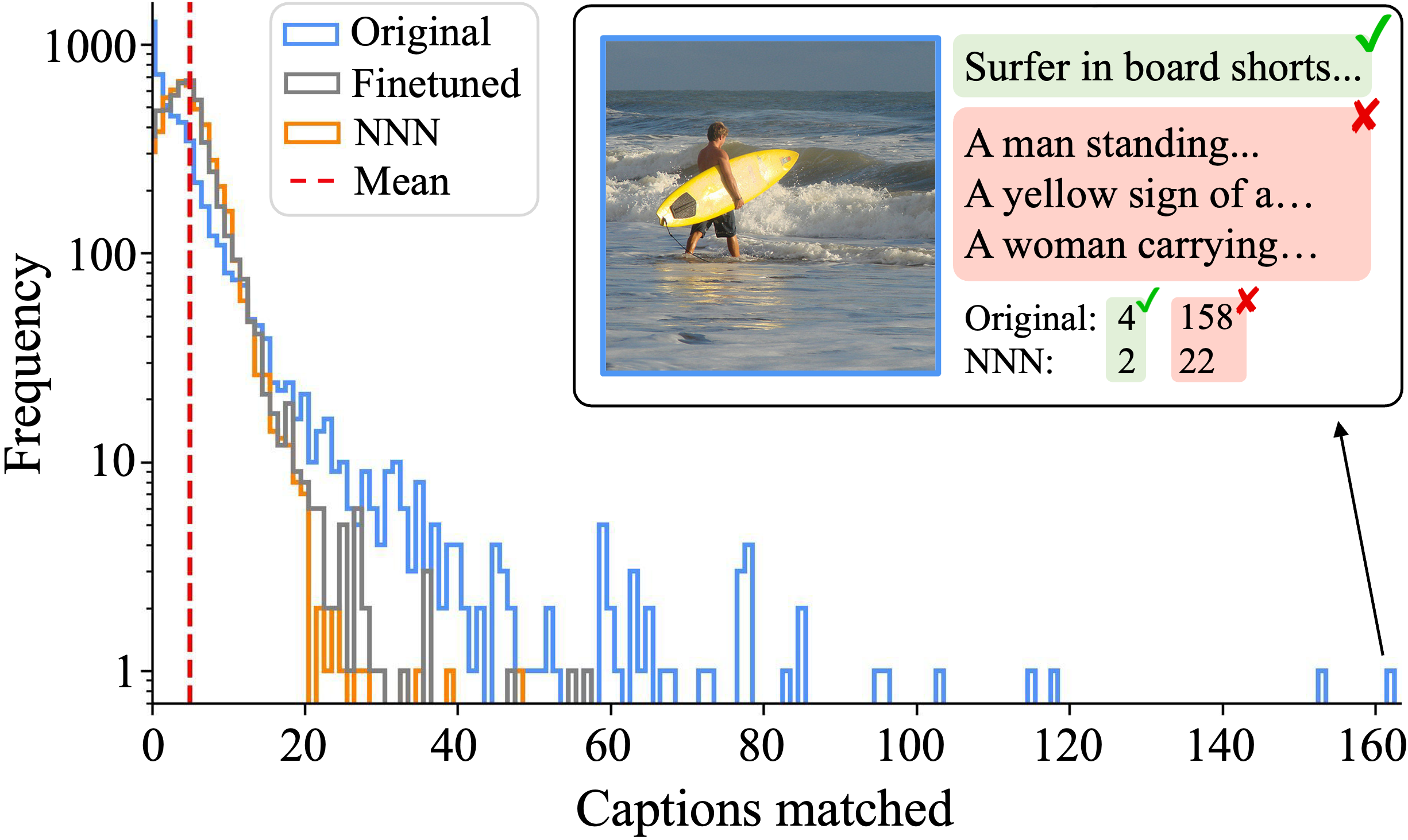}
    \vspace{-6mm}
    \caption{\textbf{Distribution of COCO captions matched to each image during image retrieval.} A base CLIP model contains many hubs that match over 100 captions, while the distribution after \nnd\ shows fewer hubs, on par with finetuning on COCO.}
    \vspace{-5mm}
    \label{fig:matching}
\end{figure}

Retrieval models compute a match score $s(q, r)$ between a query $q$ and database retrieval candidate $r$, and return the highest-scoring candidates. In the case of contrastive multimodal models such as CLIP, this score is typically the cosine similarity between image and text embeddings \citep{radford2021learning}. Figure \ref{fig:matching} shows how the hubness problem \cite{radovanovic2010hubs} manifests as a failure mode of contrastive text-to-image retrieval. Some images are simply preferred by contrastive models over other images: they have high cosine similarity with a wide array of query captions. 

To correct for bias towards hubs in image-text retrieval, we propose \nnd, an approach that estimates bias for each retrieval candidate using a database of reference queries, $\mathcal{D}$. The bias is then applied as an additive correction to the original match score, then used for retrieval. Specifically, given a contrastive retrieval score $s(q, r) = q \cdot r$, we define the bias $b(r)$ for a retrieval candidate $r$ as a constant multiple ($\alpha$) of the mean of $s(q_1, r), s(q_2, r), \dots, s(q_k, r)$, where $\{q_1, \dots, q_k\} = \mathcal{D}_{\text{top } k}(r)$ are the $k$ queries from the reference query dataset that have the highest similarity score $s(q_i, r)$ with $r$. Namely, if we define the operator $\text{argmax}^k$ to denote the $k$ arguments for the which a function attains its $k$ maximum values, then we have $D_{\text{top } k}(r) = \underset{q \in \mathcal{D}}{\arg \max^k_ s(q, r)}$, and our bias is computed as:
\begin{equation}
\label{eqn:bias}
b(r) = \alpha \cdot \frac{1}{k} \sum_{q_j \in D_{\text{top }k}(r)} s(q_j, r).\end{equation}

\nnd\ uses the nearest $k$ query embeddings to differentiate similar objects, capturing fine-grained distinctions between retrieval candidates. 

Each retrieval candidate has a constant bias score, so these scores can be computed offline and cached. 
The debiased retrieval score can then be computed by subtracting the estimated bias from the original score:
\begin{equation}
\label{eqn:debias}
s_{\mathcal{D}}(q,r) = s(q, r) - b(r)
.\end{equation}

When using vector retrieval to compute match scores, bias scores are computed in sublinear time and add a constant factor to retrieval runtime; see Section~\ref{sec:retrieval_perf} for further discussion. 

%% file: 03_experiments.tex
\section{Experiments}
\label{sec:experiments}

\begin{table*}[t!]
\centering
\resizebox{\textwidth}{!}{

\begin{tabular}{@{}lcccccccc@{}}
\toprule
& \multicolumn{4}{c}{Flickr30k retrieval} & \multicolumn{4}{c}{COCO retrieval} \\
\cmidrule(lr){2-5} \cmidrule(lr){6-9}
& Original &  DBNorm & \nnd\ Flickr & \nnd\ COCO & Original & DBNorm & \nnd\ Flickr & \nnd\ COCO \\ 
\midrule
CLIP & 58.82 & \textbf{65.26 (+6.4)} & 64.60  (+5.8)& 63.70  (+4.9)& 30.43 & \textbf{37.82 (+7.4)}& 33.45  (+3.0) & 37.53 (+7.1)\\
CLIP ft. Flickr & 72.80 & 73.80 (+1.0)& \textbf{74.14 (+1.3)} & 73.32  (+0.5) & 35.56 & \textbf{40.19 (+4.6)}& 36.25  (+0.7) & 40.12 (+4.6)\\
CLIP ft. COCO & 67.40 & 68.36 (+1.0) & \textbf{68.86 (+1.5)} & 68.04  (+0.6) & 45.89 & \textbf{47.57 (+1.7)}& 46.14  (+0.2) & 47.39  (+1.5)\\
BLIP ft. Flickr & 83.58 & 83.12 (-0.5) & \textbf{84.32 (+0.7)} & 84.06  (+0.5) & 56.44 & \textbf{59.72 (+3.3)}& 57.22  (+0.8) & 59.70 (+3.3)\\
BLIP ft. COCO & 82.12 & 81.92 (-0.2) & \textbf{82.80 (+0.7)}& 82.64  (+0.5) & 62.68 & 64.00 (+1.3)& 62.82  (+0.1) & \textbf{64.44 (+1.8)}\\
ALBEF ft. Flickr & 79.50 & 79.86 (+0.4) & \textbf{80.26 (+0.8)} & 79.90  (+0.4)& 52.53 & 56.62 (+4.1)& 53.18  (+0.6) & \textbf{56.67  (+4.1)}\\
ALBEF ft. COCO & 74.54 & 76.10 (+1.6)& \textbf{76.60 (+2.1)}& 75.80  (+1.3)& 59.73 & \textbf{62.72 (+3.0)}& 60.10  (+0.4)& 62.66 (+2.9)\\
SigLIP & 74.62 & 76.02 (+1.4) & \textbf{76.54 (+1.9)} & 76.08  (+1.5) & 47.15 & 49.93 (+2.8)& 48.49  (+1.3) & \textbf{50.24 (+3.1)}\\
BEiT-3 & 75.52 & 76.08 (+0.6) & \textbf{76.66 (+1.1)} & 76.30  (+0.8)& 47.62 & 50.08 (+2.5)& 47.93  (+0.3) & \textbf{50.64 (+3.0)}\\
BEiT-3 ft. Flickr & 86.12 & 84.68 (-1.4) & 86.00  (-0.1)& \textbf{86.30 (+0.2)} & 53.57 & 55.16 (+1.6)& 53.79  (+0.2) & \textbf{55.91 (+2.3)}\\
BEiT-3 ft. COCO & 82.90 & 82.20 (-0.7) & \textbf{83.48 (+0.6)} & 82.78  (-0.1) & 61.88 & 61.78 (-0.1)& 61.60  (-0.3)& \textbf{62.34 (+0.5)}\\
BEiT-3 Large & 77.80 & 77.70 (-0.1) & \textbf{78.54 (+0.7)} & 78.20  (+0.4)& 49.34 & 51.67 (+2.3)& 50.24  (+0.9) & \textbf{52.25 (+2.9)}\\
BEiT-3 Large ft. Flickr & \textbf{88.04} & 86.74 (-1.3) & 87.82  (-0.2) & 87.70  (-0.3)& 56.41 & 58.09 (+1.7)& 56.68  (+0.3) & \textbf{58.88 (+2.5)}\\
BEiT-3 Large ft. COCO & 86.24 & 85.12 (-1.1) & \textbf{86.64 (+0.4)} & 86.18  (-0.1) & 63.83 & 63.57 (-0.3)& 63.75  (-0.1) & \textbf{64.20 (+0.4)}\\
\bottomrule
\end{tabular}
}
\vspace{-2mm}
\caption{\textbf{Image Recall@1 results for Flickr30k and COCO.}  \% change in parantheses; ``ft.'' indicates finetuned. } 
\vspace{1mm}
\label{tab:image_retrieval_results} 
\end{table*}

\begin{table*}[t!]
\centering
\resizebox{\textwidth}{!}{

\begin{tabular}{@{}lcccccccc@{}}
\toprule
& \multicolumn{4}{c}{Flickr30k retrieval} & \multicolumn{4}{c}{COCO retrieval} \\
\cmidrule(lr){2-5} \cmidrule(lr){6-9}
& Original & DBNorm & \nnd\ Flickr & \nnd\ COCO & Original & DBNorm & \nnd\ Flickr & \nnd\ COCO \\ 
\midrule
CLIP & 79.30& \textbf{81.20 (+1.9)}& \textbf{81.20 (+1.9)}& 80.10  (+0.8)& 50.02 & 53.20  (+3.2)& 51.60  (+1.6) & \textbf{53.66 (+3.6)}\\
CLIP ft. Flickr & 85.70& 86.50  (+0.8)& \textbf{87.30 (+1.6)}& 86.60  (+0.9)& 53.74 & 55.42  (+1.7) & 53.92  (+0.2) & \textbf{56.44 (+2.7)}\\
CLIP ft. COCO & 82.10& 81.90  (-0.2)& \textbf{82.80 (+0.7)}& 82.70 (+0.6)& 63.74 & 64.72  (+1.0)& 63.88  (+0.1) & \textbf{65.26 (+1.5)}\\
BLIP ft. Flickr & 93.40& \textbf{95.70 (+2.3)}& 95.20  (+1.8)& 94.30  (+0.9)& 72.26 & 78.28 (+6.0)& 75.90  (+3.6)& \textbf{78.30  (+6.0)}\\
BLIP ft. COCO & 93.70& 94.70  (+1.0)& \textbf{95.30 (+1.6)}& 94.60  (+0.9)& 79.62 & \textbf{82.52  (+2.9)}& 79.58  (-0.0) & 82.46 (+2.8)\\
ALBEF ft. Flickr & 92.40& \textbf{93.10  (+0.7)}& 92.60  (+0.2)& 92.70 (+0.3)& 69.82 & \textbf{74.62  (+4.8)}& 71.06  (+1.2) & 74.44 (+4.6)\\
ALBEF ft. COCO & 87.30& \textbf{90.50 (+3.2)}& 90.00  (+2.7)& 89.30  (+2.0)& 78.60& 80.54  (+1.9) & 79.10  (+0.5)& \textbf{80.68 (+2.1)}\\
SigLIP & 89.00& \textbf{91.60 (+2.6)}& 91.30  (+2.3)& 91.30  (+2.3)& 65.32 & 69.14  (+3.8) & 66.80  (+1.5)& \textbf{69.86 (+4.5)}\\
BEiT-3 & 89.10& 90.70  (+1.6)& \textbf{91.80 (+2.7)}& 90.90  (+1.8)& 61.12 & 68.94  (+7.8) & 65.66  (+4.5) & \textbf{69.12 (+8.0)}\\
BEiT-3 ft. Flickr & \textbf{96.30}& 94.40  (-1.9)& 95.60  (-0.7)& 95.90  (-0.4)& 72.02 & 75.12  (+3.1) & 72.62  (+0.6) & \textbf{75.22 (+3.2)}\\
BEiT-3 ft. COCO & 93.60& 94.50  (+0.9)& \textbf{95.30 (+1.7)}& 94.80  (+1.2)& 80.72 & 79.90  (-0.8) & 80.42  (-0.3) & \textbf{81.26 (+0.5)}\\
BEiT-3 Large & 91.10& \textbf{93.20  (+2.1)}& \textbf{93.20 (+2.1)}& 92.20  (+1.1)& 63.26 & 71.06 (+7.8)& 67.60  (+4.3)& \textbf{71.08  (+7.8)}\\
BEiT-3 Large ft. Flickr & 97.20& 96.80  (-0.4)& 97.20  (0.0)& \textbf{97.50 (+0.3)}& 74.32 & 77.56  (+3.2) & 74.86  (+0.5) & \textbf{77.92 (+3.6)}\\
BEiT-3 Large ft. COCO & 95.50& 95.00  (0.0)& 95.30  (-0.2)& \textbf{96.20 (+0.7)}& 82.10& 80.88  (-1.2) & 81.98  (-0.1) & \textbf{82.72 (+0.6)}\\
\bottomrule
\end{tabular}
}
\vspace{-2mm}
\caption{\textbf{Text Recall@1 Results for Flickr30k and COCO.} \% change in parantheses; ``ft.'' indicates finetuned. } 
\vspace{-4mm}
\label{tab:text_retrieval_results} 
\end{table*}

We evaluate \nnd\ on both text-to-image and image-to-text retrieval using a variety of contrastive multimodal models (CLIP, BLIP, ALBEF, SigLIP, BEiT) \citep{radford2021learning,li2021align, zeng2021multi, li2022blip, wang2022image, zhai2023sigmoid} on well-established retrieval datasets Flickr30k and COCO \citep{Young2014FromID, lin2015microsoft}. We also report the accuracy of DBNorm, the top-performing baseline, using DBNorm's DualIS scoring function \cite{wang-etal-2023-balance}. Additional DN \cite{zhou2023testtime}, QBNorm \cite{bogolin2022cross}, and DualDIS (a similar performing variant of DualIS) baselines are discussed in Appendix \ref{sec:appendix_retrievalresults}.

\label{sec:biased_retrieval}
\subsection{Retrieval performance}
\label{sec:retrieval_perf}
\vspace{-1mm}
\paragraph{Accuracy.} To evaluate the impact of \nnd\ on retrieval performance, we hold out a random subset of the training set with the same size as the test set, and optimize $\alpha$ and $k$ via a hyperparameter search (Appendix \ref{sec:appendix_A}). We use the same approach to optimize the DBNorm hyperparameters (but we note that optimizing these parameters takes 100x the compute). Then, we evaluate both methods on the test set: for image retrieval, we use training captions as the reference database, and for text retrieval, we use training images. Full results are shown for image retrieval (Table~\ref{tab:image_retrieval_results}) and text retrieval (Table~\ref{tab:text_retrieval_results}) for Recall@1 (using 20\% of training data as the reference database, following \citet{wang-etal-2023-balance}). Appendix \ref{sec:appendix_retrievalresults} includes results and confidence intervals for Recall@5 and Recall@10.

\begin{figure*}[htbp!]
\centering
\includegraphics[width=\textwidth]{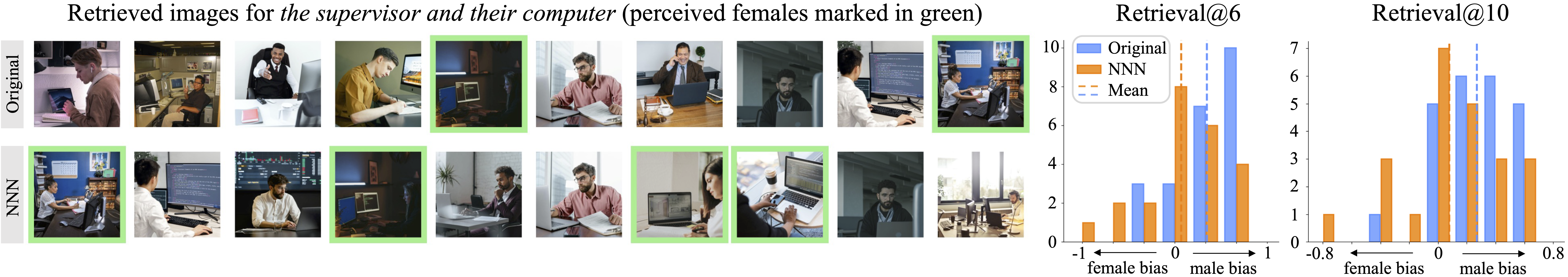}
\vspace{-7mm}
\caption{\textbf{\nnd\ decreases gender bias in image retrieval}. (L) Top 10 retrieved Visogender images for an example query, before (\textit{top}) and after (\textit{bottom}) \nnd\ debiasing. (R) Distribution of image retrieval bias across occupations.}
\label{fig:gender}
\vspace{-2mm}
\end{figure*}

We performed experiments with both in-distribution queries (e.g. 
 normalizing COCO retrieval using COCO reference queries) and out-of-distribution queries (e.g. normalizing Flickr using COCO). \nnd\ still shows consistent gains over the original model when scores are normalized with out-of-distribution queries. We also ran ablation studies on the size of the reference query database using various subsets of Flickr and COCO and find minimal performance decrease (see Appendix \ref{sec:appendix_ablation}).
\vspace{-1mm}
\paragraph{Efficiency.}

Since \nnd\ only requires the $k$-nearest reference queries per retrieval candidate, unlike QBNorm and DBNorm, it does not require an exhaustive search over the $|\textsc{retrieval dataset}| \times |\textsc{reference dataset}|$ matrix of similarity scores. We can use an inverted file index from Faiss \cite{douze2024faiss} to efficiently compute the per-retrieval candidate bias scores. Then, to use bias scores in retrieval with a vector index, we modify retrieval embedding $r$ to $r' = \langle r, b \rangle$, where $b$ is the associated bias with $r$, and modify query embedding $q$ to $q' = \langle q, -1 \rangle$. Thus, the new inner product between $r'$ and $q'$ is $r' \cdot q' = r \cdot q - b$, which is equivalent to Equation~\ref{eqn:debias}. Table~\ref{tab:gpu_search_comparison} shows that for \nnd, using a vector index for both operations causes over a 100x increase in speed over exhaustive search with only a minor performance drop (maximum $-0.2\%$ accuracy).

\subsection{Correcting image and caption bias}

To provide intuition on how \nnd\ impacts hubness, we analyzed \textit{hub} images that match with many queries, despite having only a few correct ground-truth captions. In Figure~\ref{fig:matching}, we show that for CLIP on COCO image retrieval, \nnd\ significantly reduces imbalance in this distribution and decreases the effect of hubs comparably to finetuning directly on the reference query dataset. Table~\ref{tab:kurtosis_comparison} further demonstrates that across models and datasets, \nnd\ decreases outlier metrics including kurtosis (tailedness) and mean absolute error. Distribution shifts for additional image and text retrieval settings (Appendix \ref{sec:appendix_imagecaptionbias}) show a similar trend.

\begin{table}[t!]
\centering
\resizebox{\columnwidth}{!}{
\begin{tabular}{lcccc}
\toprule
 & \multicolumn{2}{c}{\textbf{CLIP}} & \multicolumn{2}{c}{\textbf{BLIP}} \\
\cmidrule(r){2-3} \cmidrule(l){4-5}
 & COCO & Flickr & COCO & Flickr \\
\addlinespace[-3pt]
\midrule
Kurtosis & 59.8 & 9.0 & 32.1 & 3.2 \\
Kurtosis (\nnd) & \textbf{9.5} & \textbf{1.1} & \textbf{12.3} & \textbf{1.9} \\
\hline
MAE & 4.8 & 2.8 & 2.1 & 1.2 \\
MAE (\nnd) & \textbf{2.6} & \textbf{1.7} & \textbf{1.6} & \textbf{1.0} \\
\hline
Max & 162 & 39 & 59 & 15 \\
Max (\nnd) & \textbf{48} & \textbf{15} & \textbf{32} & \textbf{12} \\
\hline
$\Delta$ accuracy & +7.4 & +6.5 & +1.8 & +1.2 \\
\addlinespace[-3pt] 
\bottomrule
\end{tabular}
}
\caption{\textbf{Outlier reduction on text-to-image retrieval.} \nnd\ leads to tighter distributions of captions retrieved per image and decreases the number of hub images.}
\label{tab:kurtosis_comparison}
\vspace{-4mm}
\end{table}

\subsection{Reducing gender bias in image retrieval} 
In addition to broad retrieval experiments, we also measure the effect of \nnd\ on unwanted correlations between specific input attributes and retrieval scores. We examine gender bias, where most corrective methods show a tradeoff between bias and retrieval accuracy: stronger debiasing is accompanied by a performance drop \citep{wang2021genderneutral, berg2022prompt, wang2022fairclip}. 
 \nnd\ reduces gender bias while \textit{improving} retrieval accuracy.

We evaluate \nnd\ on CLIP for a subset of the VisoGender benchmark \citep{hall2023visogender}, which contains images of people and objects corresponding to 23 occupations (5 images perceived male and 5 female per occupation), and associated gender-neutral captions of the form ``The \textit{occupation} and their \textit{object}.'' Retrieval returns the closest $n$ images for a caption (\eg\ the \textit{supervisor} and their \textit{computer}). Applying \nnd\ to this setting requires a choice of reference captions, as VisoGender does not include a training distribution.
Experiments using the COCO training set (with hyperparameters from Table \ref{tab:hyperparams_text_to_image}, $k=16$, $\alpha=0.75$)
found significant decreases in mean gender bias on VisoGender image retrieval. These results demonstrate the flexibility of \nnd\ for settings without an obvious reference database. Further work could also explore generation of task-specific reference sets.

An example of our method successfully debiasing images retrieved for an input query is shown in Figure \ref{fig:gender}. We also plot the distribution of the bias ($\frac{\text{\# men} - \text{\# women}}{n}$) across all the occupations at $n=6,10$. While the original CLIP retrieval results are significantly biased towards men, \nnd\ shifts the average bias toward 0 (reduces from 0.348 to 0.072 for $n=6$, and from 0.270 to 0.078 for $n=10$).

Importantly, we find that \nnd\ simultaneously boosts average precision (the proportion of retrieved images matching the occupation described in the caption) from $56.5\%$ to $69.6\%$ (Retrieval@1) and from $49.6\%$ to $56.5\%$ (Retrieval@5).

%% file: 04_conclusion.tex
\vspace{2mm}
\section{Conclusion}
\vspace{2mm}
\label{sec:conclusion}

We introduce Nearest Neighbor Normalization for contrastive multimodal retrieval. By precomputing bias correction scores using only the \textit{k}-nearest neighbors, \nnd\ is substantially more efficient while slightly improving accuracy over previous test-time inference methods. We also show that \nnd\ can be used flexibly with arbitrary reference datasets and performs well at reducing gender bias. 

%% file: 045_ethics_limitations.tex
\section{Limitations}

\nnd\ can be applied to contrastive multimodal models to achieve significant and consistent retrieval score improvements. We have not shown that the same holds for models with a dedicated cross-attention between image and text embeddings, and show evidence that it might not be effective in Appendix~\ref{sec:appendix_crossmodal}. Furthermore, although \nnd\ is fast for contrastive models due to the efficiency of vector retrieval, it is much slower for crossmodal models, as computing each image-text matching score requires a forward pass.

\section{Ethical considerations}

Contrastive models can be used in consumer-facing retrieval and search systems by major tech companies, and so failures can have a wide impact. Extensive bias has been documented in such models \cite{wang2021genderneutral, wang2022fairclip, berg2022prompt}. Although our paper primarily evaluates the generic case of improving multimodal retrieval scores, we have also shown that \nnd\ works to debias targeted attributes, such as gender. Still, our method should not be seen as a replacement for human oversight and careful training dataset curation. 

\section{Acknowledgements}
We are grateful for the support of the MIT-IBM Watson AI Lab and ARL grant W911NF-18-2-0218. We are grateful to teaching staff of the MIT 6.8611 Quantitative Methods in Natural Language class, where many of the authors began their work on this project. We also thank Ethan Chang and Tazo Chowdhury for ongoing support.

%% file: 05_appendix.tex
\setcounter{table}{0}
\renewcommand{\thetable}{A\arabic{table}} 
\setcounter{subsection}{0}
\renewcommand{\thesubsection}{A\arabic{subsection}}

\setcounter{figure}{0}
\renewcommand{\thefigure}{A\arabic{figure}} 

\setcounter{section}{0}
\setcounter{subsection}{0}
\renewcommand{\thesubsection}{\Alph{section}\arabic{subsection}} 

\section*{Appendix}
\label{sec:appendix}

\section{Baselines}
\subsection{DBNorm}

The main DBNorm scoring function, DualIS \cite{wang-etal-2023-balance}, is described as follows: given a query $q$, retrieval candidate $r_i$, reference query database $\hat{Q}$, and reference retrieval candidate database $\hat{R}$, the normalized score $\hat{s}(q, r_i)$ is computed using the following expressions (where $s(q, r)$ denotes the dot product score between the embeddings):

\begin{equation}
    \hat{s}({q, r_i}) = \hat{s}^{\hat{R}}_{q, r_i} * \hat{s}^{\hat{Q}}_{q, r_i}
\end{equation}

\begin{equation}
    \hat{s}^{\hat{R}}_{q, r_i} = \frac{\exp(\beta_1 s(q, r_i))}{\sum_{\hat{r} \in \hat{R}} \exp(\beta_1 s(\hat{r}, r_i))}
\end{equation}

\begin{equation}
    \hat{s}^{\hat{Q}}_{q, r_i} = \frac{\exp(\beta_2 s(q, r_i))}{\sum_{\hat{q} \in \hat{Q}} \exp(\beta_2 s(\hat{q}, r_i))}
\end{equation}
DualDIS is a variant of DualIS that uses the original $s(q, r_i)$ score instead of $\hat{s}^{\hat{R}}_{q, r_i}$ or $\hat{s}^{\hat{Q}}_{q, r_i}$ for a given query $q$ if the closest retrieval candidate to $q$ is not in a precomputed ``activation set'' that contains all likely hubs. See \citet{wang-etal-2023-balance} for details on how the activation sets are computed. In our experiments, we find that DualDIS and DualIS are very similar in performance (Table~\ref{tab:image_retrieval_results_full}, \ref{tab:text_retrieval_results_full}).

In our experiments, we use the training images as the reference retrieval candidate database for image retrieval and the training captions for text retrieval. Note that \nnd\ has the advantage of requiring a reference query database only, and does not use a reference retrieval candidate database. Moreover, \nnd\ has a constant runtime with respect to the reference database size for calculating each individual normalized score while DBNorm has a linear runtime since the summation in the denominator requires all reference embeddings.


\subsection{QBNorm}

QBNorm \cite{bogolin2022cross} is equivalent to DBNorm when $\beta_1$ is set to 0. Since our hyperparameter sweep of DBNorm includes $\beta_1 = 0$, we do not explicitly include QBNorm as a baseline in our results.

\subsection{Distribution Normalization (DN)}

DN \cite{zhou2023testtime} computes a first-order approximation of the DualIS normalization score by normalizing the query and retrieval embeddings to have zero mean based on reference datasets. While it also has constant time performance for each query, we find that it has far lower accuracy gains than \nnd.

\subsection{Results for all methods}

A full comparison of DN, DualIS, DualDIS, and \nnd\ is shown in Table~\ref{tab:image_retrieval_results_full} and \ref{tab:text_retrieval_results_full}.

\section{Hyperparameter selection}
\subsection{\nnd}
\label{sec:appendix_A}
We compute the hyperparameters used for retrieval in Section \ref{sec:biased_retrieval} on a per-model, evaluation dataset, and reference query dataset basis. To do so, we perform a hyperparameter sweep on
\[
\alpha \in \{0.25, 0.375, 0.5, \dots, 1.5\}
\]
and
\[
k \in  \{1, 2, 4, \dots, 512\}.
\]
We evaluate hyperparameters with image retrieval performed on a randomly selected split of the training set from the evaluation dataset. For Flickr30k, we take a split of 1,000 images and their 5,000 corresponding captions, and for COCO, we take a split of 5,000 images and their 25,000 corresponding captions. When selecting hyperparameters, we optimize for R@1 accuracy, and find that this generally does not come with significant degredation in R@5 or R@10 performance. We present the hyperparameters we use for text-to-image retrieval in Table~\ref{tab:hyperparams_text_to_image} and for image-to-text retrieval in Table~\ref{tab:hyperparams_image_to_text}.

\begin{table}[ht]
\centering
\resizebox{\columnwidth}{!}{
\begin{tabular}{lcccc} 
\toprule
& \multicolumn{2}{c}{Flickr30k, \nnd\ w/} & \multicolumn{2}{c}{COCO, \nnd\ w/}\\
& Flickr30k & COCO & Flickr30k & COCO \\
\midrule
CLIP & (0.75, 128) & (0.75, 16) & (0.5, 8) & (0.75, 256)\\
CLIP ft. Flickr & (0.5, 32) & (0.25, 128) & (0.5, 32) & (0.75, 256)\\
CLIP ft. COCO & (0.5, 16) & (0.5, 1) & (0.25, 16) & (0.75, 128)\\
BLIP & (0.5, 16) & (0.25, 4) & (0.25, 4) & (0.75, 64)\\
BLIP ft. Flickr & (0.5, 32) & (0.25, 4) & (0.5, 64) & (0.75, 16)\\
ALBEF ft. Flickr & (0.75, 32) & (0.25, 16) & (0.5, 4) & (0.75, 256)\\
ALBEF ft. COCO & (0.75, 32) & (0.5, 16) & (0.25, 8) & (0.75, 128)\\
SigLIP & (0.75, 128) & (0.5, 128) & (0.5, 16) & (0.75, 128)\\
BEiT-3 & (0.75, 32) & (0.5, 64) & (0.25, 4) & (0.75, 128)\\
BEiT-3 ft. Flickr & (0.25, 8) & (0.25, 64) & (0.25, 4) & (0.75, 256)\\
BEiT-3 ft. COCO & (0.75, 16) & (0.25, 2) & (0.25, 32) & (0.25, 128)\\
BEiT-3 Large & (0.5, 256) & (0.5, 32) & (0.25, 32) & (0.75, 128)\\
BEiT-3 Large ft. Flickr & (0.5, 16) & (0.25, 1) & (0.25, 16) & (0.75, 512)\\
BEiT-3 Large ft. COCO & (0.5, 8) & (0.25, 128) & (0.25, 8) & (0.5, 64)\\
\bottomrule
\end{tabular}
}
\vspace{1mm}
\caption{\textbf{Optimal $(\alpha, k)$ for model, evaluation, and reference query dataset triples for text-to-image retrieval.}} 
\label{tab:hyperparams_text_to_image} 

\end{table}

\begin{table}[ht]
\centering
\resizebox{\columnwidth}{!}{
\begin{tabular}{lcccc} 
\toprule
& \multicolumn{2}{c}{Flickr30k, \nnd\ w/} & \multicolumn{2}{c}{COCO, \nnd\ w/}\\
& Flickr30k & COCO & Flickr30k & COCO \\
\midrule
CLIP & (0.75, 16) & (0.5, 2) & (0.5, 8) & (0.75, 128)\\
CLIP ft. Flickr & (0.5, 16) & (0.25, 1) & (0.25, 2) & (0.5, 128)\\
CLIP ft. COCO & (0.5, 32) & (0.25, 16) & (0.25, 16) & (0.75, 64)\\
BLIP & (1, 512) & (0.75, 16) & (0.5, 16) & (0.75, 32)\\
BLIP ft. Flickr & (0.75, 512) & (0.75, 64) & (0.75, 32) & (0.75, 64)\\
ALBEF ft. Flickr & (0.25, 512) & (0.25, 64) & (0.5, 16) & (0.75, 128)\\
ALBEF ft. COCO & (0.75, 32) & (0.5, 64) & (0.25, 8) & (0.75, 32)\\
SigLIP & (0.5, 64) & (0.75, 256) & (0.25, 32) & (0.75, 128)\\
BEiT-3 & (0.75, 64) & (0.5, 32) & (0.5, 32) & (0.75, 256)\\
BEiT-3 ft. Flickr & (1, 32) & (0.75, 4) & (0.25, 16) & (0.75, 256)\\
BEiT-3 ft. COCO & (0.5, 32) & (0.5, 4) & (0.25, 4) & (0.5, 8)\\
BEiT-3 Large & (0.5, 64) & (0.5, 512) & (0.5, 16) & (0.75, 512)\\
BEiT-3 Large ft. Flickr & (0.5, 64) & (0.75, 16) & (0.5, 16) & (0.75, 128)\\
BEiT-3 Large ft. COCO & (0.5, 64) & (0.75, 32) & (0.25, 64) & (0.5, 16)\\
\bottomrule
\end{tabular}
}
\vspace{1mm}
\caption{\textbf{Optimal $(\alpha, k)$ for model, evaluation, and reference query dataset triples for image-to-text retrieval.}} 
\label{tab:hyperparams_image_to_text} 

\end{table}

We find four main trends in hyperparameter selection: (1) for out-of-distribution reference query databases, smaller $\alpha$ (0.25 to 0.5) and $k$ (8 to 16) are optimal, and for in-distribution reference query sets, larger $\alpha$ (0.75) are optimal; (2) model and dataset pairs with higher baseline retrieval scores see greater improvements from small $\alpha$ and $k$; (3) hyperparameters transfer well across text-to-image and image-to-text retrieval; (4) for in-distribution reference query sets with $\alpha = 0.75$, our method is not very sensitive to choice of $k$. We see improvements from $k$ even as small as 1 to 8, and similar improvements for $k$ ranging from 8 to 128, as shown in Tables \ref{tab:k_sweep_text_to_image} (for image retrieval) and \ref{tab:k_sweep_image_to_text} (for text retrieval).

\begin{table}[ht]
\centering
\resizebox{\columnwidth}{!}{
\begin{tabular}{lcccccccc} 
\toprule
& Original & $k =$ 1 & 4 & 8 & 16 & 32 & 64 & 128 \\
\midrule
CLIP & 30.45 & 35.47 & 36.57 & 36.96 & 37.36 & 37.52 & 37.67 & 37.77 \\ 
BLIP ft. COCO & 62.72 & 63.42 & 64.12 & 64.22 & 64.38 & 64.35 & 64.49 & 64.46 \\ 
CLIP ft. COCO & 45.92 & 45.08 & 46.4 & 46.88 & 47.29 & 47.51 & 47.73 & 47.93 \\ 
CLIP ft. Flickr & 35.58 & 37.75 & 38.44 & 38.91 & 39.21 & 39.61 & 40.01 & 40.16 \\ 
BLIP ft. Flickr & 56.47 & 58.94 & 59.72 & 59.92 & 60.03 & 60.04 & 60.16 & 60.22 \\ 
SigLIP & 47.18 & 48.54 & 49.5 & 49.9 & 50.23 & 50.45 & 50.6 & 50.72 \\ 
ALBEF ft. Flickr & 52.56 & 55.22 & 56.34 & 56.57 & 56.88 & 57.07 & 57.12 & 57.12 \\ 
ALBEF ft. COCO & 59.76 & 60.93 & 61.9 & 62.23 & 62.47 & 62.69 & 62.9 & 62.92 \\ 
BEiT-3 & 47.64 & 49.42 & 50.25 & 50.58 & 50.84 & 50.88 & 50.89 & 50.83 \\ 
BEiT-3 ft. Flickr & 53.59 & 54.36 & 55.3 & 55.61 & 55.99 & 56.15 & 56.28 & 56.32 \\ 
BEiT-3 ft. COCO & 61.91 & 60.52 & 61.54 & 61.86 & 62.18 & 62.46 & 62.57 & 62.61 \\ 
BEiT-3 Large & 49.36 & 51.2 & 51.91 & 52.24 & 52.46 & 52.51 & 52.52 & 52.54 \\ 
BEiT-3 Large ft. Flickr & 56.43 & 57.35 & 58.38 & 58.54 & 58.66 & 58.78 & 58.96 & 59.04 \\ 
BEiT-3 Large ft. COCO & 63.85 & 62.5 & 63.3 & 63.77 & 64.01 & 64.17 & 64.27 & 64.41 \\ 
\bottomrule
\end{tabular}
}
\vspace{1mm}
\caption{\textbf{Image Recall@1 for COCO with \nnd\ across different $k$, with fixed $\alpha = 0.75$.}} 
\label{tab:k_sweep_text_to_image} 

\end{table}

\begin{table}[ht]
\centering
\resizebox{\columnwidth}{!}{
\begin{tabular}{lcccccccc} 
\toprule
& Original & $k =$ 1 & 4 & 8 & 16 & 32 & 64 & 128 \\
\midrule
CLIP & 50.02 & 50.04 & 52.14 & 52.56 & 52.96 & 53.5 & 53.94 & 54.16 \\
BLIP ft. COCO & 79.62 & 80.56 & 81.68 & 82.32 & 82.74 & 82.68 & 82.7 & 82.46 \\
CLIP ft. COCO & 63.74 & 60.68 & 62.9 & 63.96 & 64.38 & 65.18 & 65.44 & 65.44 \\
CLIP ft. Flickr & 53.74 & 52.74 & 54.68 & 55.66 & 56.3 & 56.64 & 56.96 & 56.28 \\
BLIP ft. Flickr & 72.26 & 76.58 & 77.96 & 78.54 & 78.36 & 78.44 & 78.64 & 78.44 \\
SigLIP & 65.32 & 65.72 & 68.22 & 68.78 & 69.4 & 69.88 & 69.98 & 70.24 \\
ALBEF ft. Flickr & 69.82 & 72.28 & 74.0 & 74.34 & 74.94 & 75.16 & 74.82 & 74.82 \\
ALBEF ft. COCO & 78.6 & 77.96 & 79.82 & 79.96 & 80.22 & 80.86 & 81.22 & 81.14 \\
BEiT-3 & 61.12 & 64.9 & 66.3 & 67.5 & 68.36 & 68.78 & 69.14 & 69.26 \\
BEiT-3 ft. Flickr & 72.02 & 72.74 & 74.22 & 74.58 & 75.1 & 75.22 & 75.56 & 75.42 \\
BEiT-3 ft. COCO & 80.72 & 77.8 & 79.72 & 80.42 & 80.9 & 81.24 & 81.14 & 81.3 \\
BEiT-3 Large & 63.26 & 66.78 & 68.38 & 69.54 & 70.32 & 70.78 & 71.24 & 71.44 \\
BEiT-3 Large ft. Flickr & 74.32 & 75.32 & 76.64 & 77.38 & 78.02 & 78.66 & 78.64 & 78.72 \\
BEiT-3 Large ft. COCO & 82.1 & 79.56 & 81.46 & 82.22 & 82.74 & 83.0 & 83.04 & 83.04 \\
\bottomrule
\end{tabular}
}
\vspace{1mm}
\caption{\textbf{Text Recall@1 for COCO with \nnd\ across different $k$, with fixed $\alpha = 0.75$.}} 
\label{tab:k_sweep_image_to_text} 

\end{table}

\subsection{DBNorm}

To tune the hyperparameters $\beta_1$ and $\beta_2$, we first performed a grid sweep in logspace on
\[\log \beta_1, \log \beta_2 \in \{\log 0.001, \ldots, \log 400\}\]
with a resolution of $20$ values. We found that the best performing $\beta_1$ and $\beta_2$ occupied a tight range, so we performed a denser sweep on 

\[\log \beta_1 \in \{\log 0.001, \ldots, \log 15\}\]
\[\log \beta_2 \in \{\log 25, \ldots, \log 200\}\] again with a resolution of $20$ values. We also test setting $\beta_1$ and $\beta_2$ to 0. To select the hyperparameters from the sweep, we use the same procedure as \nnd.

\section {Runtime}
\label{sec:appendix_runtime}

A quantitative comparison of NNN runtimes using an exhaustive search (``Base'' column) on GPU and using a Faiss index for computing bias scores is shown in  Table~\ref{tab:gpu_search_comparison}. All of our experiments can be run using a single NVIDIA V100 GPU. 

\begin{table}[ht]
\centering
\resizebox{\columnwidth}{!}{
\begin{tabular}{lcccccc}
\toprule
Model & Base (s) & Faiss (s) & Factor & Base IR@1 & Faiss IR@1 \\
\midrule
CLIP & 22.69 s & 0.41 s & 55.26x & 37.76 & 37.67 \\
CLIP ft. Flickr & 20.95 s & 0.13 s & 161.4x & 40.36 & 40.33 \\
CLIP ft. COCO & 20.94 s & 0.15 s & 138.18x & 47.93 & 47.81 \\
BLIP ft. Flickr & 10.58 s & 0.07 s & 159.24x & 60.03 & 59.97 \\
BLIP ft. COCO & 10.59 s & 0.16 s & 65.07x & 64.49 & 64.45 \\
ALBEF ft. Flickr & 10.61 s & 0.07 s & 147.48x & 56.89 & 56.80 \\
ALBEF ft. COCO & 10.59 s & 0.07 s & 150.79x & 62.92 & 62.82 \\
SigLIP & 31.25 s & 0.21 s & 151.33x & 50.72 & 50.52 \\
\bottomrule
\end{tabular}
}
\vspace{1mm}
\caption{\textbf{GPU-based exhaustive search vs GPU-based vector index search for computing bias scores on COCO.}} 
\label{tab:gpu_search_comparison} 
\end{table}


\section{Full retrieval results}
\label{sec:appendix_retrievalresults}
We present the full results of \nnd\ applied to both text-to-image and image-to-text retrieval for the Flickr30k and COCO datasets, including R@1, 5, and 10 with associated 95\% confidence intervals in tables \ref{tab:flickr30k_image_retrieval_results}, \ref{tab:flickr30k_text_retrieval_results}, \ref{tab:coco_image_retrieval_results}, \ref{tab:coco_text_retrieval_results}. \nnd\  provides a consistent improvement in performance, even at higher recall values, but provides the greatest improvement to R@1. Confidence intervals are computed with bootstrapping.

\begin{table*}[t!]
\centering
\resizebox{\textwidth}{!}{

\begin{tabular}{@{}lcccccccccc@{}}
\toprule
& \multicolumn{5}{c}{Flickr30k retrieval} & \multicolumn{5}{c}{COCO retrieval} \\
\cmidrule(lr){2-6} \cmidrule(lr){7-11}
& Original & DN & DualIS & DualDIS & \nnd & Originl & DN & DualIS & DualDIS & \nnd \\ 
\midrule
CLIP & 58.82 & 62.06 & \textbf{65.26}& 65.20& 64.60 & 30.43 & 32.47 & \textbf{37.82}& 37.81& 37.53\\
CLIP ft. Flickr & 72.80 & 70.92 & 73.80& 73.78& \textbf{74.14} & 35.56 & 35.52 & \textbf{40.19}& 40.17& 40.12\\
CLIP ft. COCO & 67.40 & 66.32 & 68.36& 68.36& \textbf{68.86} & 45.89 & 45.02 & \textbf{47.57}& 47.60& 47.39\\
BLIP ft. Flickr & 83.58 & 83.74 & 83.12& 83.14& \textbf{84.32} & 56.44 & 58.15 & 59.72& \textbf{59.73}& 59.70\\
BLIP ft. COCO & 82.12 & 81.52 & 81.92& 81.92& \textbf{82.80} & 62.68 & 62.95 & 64.00& 64.00& \textbf{64.44}\\
ALBEF ft. Flickr & 79.50 & 79.18 & 79.86& 79.86& \textbf{80.26} & 52.53 & 53.92 & 56.62& \textbf{56.70}& 56.67\\
ALBEF ft. COCO & 74.54 & 74.50 & 76.10& 76.10& \textbf{76.60}& 59.73 & 60.63 & \textbf{62.72}& 62.66& 62.66\\
SigLIP & 74.62 & 75.22 & 76.02& 76.04& \textbf{76.54} & 47.15 & 47.75 & 49.93& 49.92& \textbf{50.24}\\
BEiT-3 & 75.52 & 75.72 & 76.08& 76.10& \textbf{76.66} & 47.62 & 47.75 & 50.08& 50.04& \textbf{50.64}\\
BEiT-3 ft. Flickr & \textbf{86.12} & 85.72 & 84.68& 84.68& 86.00 & 53.57 & 53.44 & 55.16& 55.16& \textbf{55.91}\\
BEiT-3 ft. COCO & 82.90 & 82.50 & 82.20& 82.20& \textbf{83.48} & 61.88 & 61.66 & 61.78& 61.78& \textbf{62.34}\\
BEiT-3 Large & 77.80 & 78.04 & 77.70& 77.74& \textbf{78.54} & 49.34 & 49.64 & 51.67& 51.70& \textbf{52.25}\\
BEiT-3 Large ft. Flickr & \textbf{88.04} & 87.40 & 86.74& 86.74& 87.82 & 56.41 & 56.82 & 58.09& 57.92& \textbf{58.88}\\
BEiT-3 Large ft. COCO & 86.24 & 85.96 & 85.12& 85.12& \textbf{86.64} & 63.83 & 63.66 & 63.57& 63.65& \textbf{64.20}\\
\bottomrule
\end{tabular}
}
\vspace{-2mm}
\caption{\textbf{Image Recall@1 results for Flickr30k and COCO.} Percent change reported for DN, DBNorm and \nnd. All methods use 20\% of the train set.} 
\vspace{1mm}
\label{tab:image_retrieval_results_full} 
\end{table*}

\begin{table*}[t!]
\centering
\resizebox{\textwidth}{!}{

\begin{tabular}{@{}lcccccccccc@{}}
\toprule
& \multicolumn{5}{c}{Flickr30k retrieval} & \multicolumn{5}{c}{COCO retrieval} \\
\cmidrule(lr){2-6} \cmidrule(lr){7-11}
& Original & DN & DualIS & DualDIS & \nnd & Original & DN & DualIS & DualDIS & \nnd \\ 
\midrule
CLIP & 79.30 & 78.50 & \textbf{81.20}& 81.10& \textbf{81.20} & 50.02 & 50.00 & 53.20& 52.92& \textbf{53.66}\\
CLIP ft. Flickr & 85.70 & 86.30 & 86.50& 86.50& \textbf{87.30} & 53.74 & 53.26 & 55.42& 55.04& \textbf{56.44}\\
CLIP ft. COCO & 82.10 & 80.80 & 81.90& 81.30& \textbf{82.80} & 63.74 & 61.80 & 64.72& 64.80& \textbf{65.26}\\
BLIP ft. Flickr & 93.40 & 95.60& \textbf{95.70}& 94.50& 95.20 & 72.26 & 75.48 & 78.28& 77.44& \textbf{78.30}\\
BLIP ft. COCO & 93.70 & 94.70 & 94.70& 94.70 & \textbf{95.30} & 79.62 & 80.30 & \textbf{82.52}& 81.72& 82.46\\
ALBEF ft. Flickr & 92.40 & 91.40 & \textbf{93.10}& 92.90& 92.60& 69.82 & 69.88 & \textbf{74.62}& 73.56& 74.44\\
ALBEF ft. COCO & 87.30 & 88.50 & \textbf{90.50}& 89.90& 90.00 & 78.60 & 78.56 & 80.54& 80.32& \textbf{80.68}\\
SigLIP & 89.00 & 89.80 & \textbf{91.60}& 91.20& 91.30 & 65.32 & 66.04 & 69.14& 69.18& \textbf{69.86}\\
BEiT-3 & 89.10 & 90.10 & 90.70 & 91.00 & \textbf{91.80} & 61.12 & 65.62 & 68.94& 68.36& \textbf{69.12}\\
BEiT-3 ft. Flickr & \textbf{96.30} & 95.30 & 94.40& 95.10& 95.60 & 72.02 & 72.96 & 75.12& 74.02& \textbf{75.22}\\
BEiT-3 ft. COCO & 93.60 & 93.90 & 94.50& 92.90& \textbf{95.30} & 80.72 & 80.14 & 79.90& 79.56& \textbf{81.26}\\
BEiT-3 Large & 91.10 & 92.70 & 93.20& \textbf{93.30}& 93.20& 63.26 & 67.20 & 71.06& 70.48& \textbf{71.08}\\
BEiT-3 Large ft. Flickr & \textbf{97.20} & 97.00 & 96.80& 96.30& \textbf{97.20} & 74.32 & 75.64 & 77.56& 76.56& \textbf{77.92}\\
BEiT-3 Large ft. COCO & 95.50 & \textbf{96.10} & 95.00& 95.10& 95.30 & 82.10 & 82.14 & 80.88& 82.32& \textbf{82.72}\\
\bottomrule
\end{tabular}
}
\vspace{-2mm}
\caption{\textbf{Text Recall@1 results for Flickr30k and COCO.} Percent change reported for DN, DBNorm and \nnd. All methods use 20\% of the train set.} 
\vspace{1mm}
\label{tab:text_retrieval_results_full} 
\end{table*}

\begin{table*}[ht]
\centering
\resizebox{16cm}{!}{
\begin{tabular}{l|ccc|ccc|ccc} 

\toprule
& \multicolumn{3}{c|}{Flickr} & \multicolumn{3}{c|}{Flickr, \nnd\ w/ Flickr} & \multicolumn{3}{c}{Flickr, \nnd\ w/ COCO}\\ 
& R@1 & R@5 & R@10 & R@1 & R@5 & R@10 & R@1 & R@5 & R@10 \\ 
\midrule
CLIP & 58.82 $\pm$ 1.36 & 83.44 $\pm$ 1.03 & 90.08 $\pm$ 0.83 & \textbf{65.52} $\pm$ \textbf{1.32} & \textbf{87.84} $\pm$ \textbf{0.91} & \textbf{93.00} $\pm$ \textbf{0.71} & 64.42 $\pm$ 1.33 & 87.24 $\pm$ 0.92 & 92.36 $\pm$ 0.74\\
CLIP ft. Flickr & 72.80 $\pm$ 1.23 & \textbf{92.54} $\pm$ \textbf{0.73} & 95.64 $\pm$ 0.57 & \textbf{74.26} $\pm$ \textbf{1.21} & 92.44 $\pm$ 0.73 & \textbf{96.22} $\pm$ \textbf{0.53} & 73.58 $\pm$ 1.22 & 92.24 $\pm$ 0.74 & 95.78 $\pm$ 0.56\\
CLIP ft. COCO & 67.40 $\pm$ 1.30 & 88.46 $\pm$ 0.89 & 93.76 $\pm$ 0.67 & \textbf{69.48} $\pm$ \textbf{1.28} & \textbf{89.64} $\pm$ \textbf{0.84} & \textbf{94.40} $\pm$ \textbf{0.64} & 67.60 $\pm$ 1.30 & 89.16 $\pm$ 0.86 & 93.84 $\pm$ 0.67\\
BLIP & 82.12 $\pm$ 1.06 & 96.10 $\pm$ 0.54 & 97.78 $\pm$ 0.41 & \textbf{83.34} $\pm$ \textbf{1.03} & \textbf{96.46} $\pm$ \textbf{0.51} & 97.90 $\pm$ 0.40 & 82.60 $\pm$ 1.05 & 96.26 $\pm$ 0.53 & \textbf{97.98} $\pm$ \textbf{0.39}\\
BLIP ft. Flickr & 83.58 $\pm$ 1.03 & 96.60 $\pm$ 0.50 & \textbf{98.50} $\pm$ \textbf{0.34} & \textbf{84.80} $\pm$ \textbf{1.00} & \textbf{96.96} $\pm$ \textbf{0.48} & 98.44 $\pm$ 0.34 & 84.22 $\pm$ 1.01 & 96.76 $\pm$ 0.49 & 98.40 $\pm$ 0.35\\
ALBEF ft. Flickr & 79.50 $\pm$ 1.12 & 95.20 $\pm$ 0.59 & 97.62 $\pm$ 0.42 & \textbf{80.84} $\pm$ \textbf{1.09} & \textbf{95.50} $\pm$ \textbf{0.57} & \textbf{97.70} $\pm$ \textbf{0.42} & 80.02 $\pm$ 1.11 & 95.44 $\pm$ 0.58 & 97.64 $\pm$ 0.42\\
ALBEF ft. COCO & 74.54 $\pm$ 1.21 & 93.32 $\pm$ 0.69 & 96.64 $\pm$ 0.50 & \textbf{76.94} $\pm$ \textbf{1.17} & \textbf{93.92} $\pm$ \textbf{0.66} & \textbf{96.90} $\pm$ \textbf{0.48} & 76.20 $\pm$ 1.18 & 93.84 $\pm$ 0.67 & \textbf{96.90} $\pm$ \textbf{0.48}\\
SigLIP & 74.62 $\pm$ 1.21 & 92.30 $\pm$ 0.74 & 95.62 $\pm$ 0.57 & \textbf{76.80} $\pm$ \textbf{1.17} & \textbf{93.30} $\pm$ \textbf{0.69} & \textbf{96.12} $\pm$ \textbf{0.54} & 76.22 $\pm$ 1.18 & 92.88 $\pm$ 0.71 & 95.84 $\pm$ 0.55\\
BEiT-3 & 75.52 $\pm$ 1.19 & 92.76 $\pm$ 0.72 & 95.96 $\pm$ 0.55 & \textbf{77.20} $\pm$ \textbf{1.16} & \textbf{93.92} $\pm$ \textbf{0.66} & \textbf{96.60} $\pm$ \textbf{0.50} & 76.36 $\pm$ 1.18 & 93.44 $\pm$ 0.69 & 96.48 $\pm$ 0.51\\
BEiT-3 ft. Flickr & 86.12 $\pm$ 0.96 & 97.68 $\pm$ 0.42 & 98.82 $\pm$ 0.30 & \textbf{86.40} $\pm$ \textbf{0.95} & \textbf{97.84} $\pm$ \textbf{0.40} & \textbf{98.88} $\pm$ \textbf{0.29} & 86.20 $\pm$ 0.96 & 97.62 $\pm$ 0.42 & 98.84 $\pm$ 0.30\\
BEiT-3 ft. COCO & 82.90 $\pm$ 1.04 & 96.54 $\pm$ 0.51 & 98.46 $\pm$ 0.34 & \textbf{83.44} $\pm$ \textbf{1.03} & \textbf{96.84} $\pm$ \textbf{0.48} & \textbf{98.62} $\pm$ \textbf{0.32} & 83.12 $\pm$ 1.04 & 96.62 $\pm$ 0.50 & 98.48 $\pm$ 0.34\\
BEiT-3 Large & 77.80 $\pm$ 1.15 & 93.92 $\pm$ 0.66 & 96.58 $\pm$ 0.50 & \textbf{78.92} $\pm$ \textbf{1.13} & \textbf{94.54} $\pm$ \textbf{0.63} & \textbf{97.14} $\pm$ \textbf{0.46} & 78.84 $\pm$ 1.13 & \textbf{94.54} $\pm$ \textbf{0.63} & 96.82 $\pm$ 0.49\\
BEiT-3 Large ft. Flickr & \textbf{88.04} $\pm$ \textbf{0.90} & 98.06 $\pm$ 0.38 & \textbf{99.04} $\pm$ \textbf{0.27} & 87.90 $\pm$ 0.90 & \textbf{98.08} $\pm$ \textbf{0.38} & 98.96 $\pm$ 0.28 & 87.82 $\pm$ 0.91 & 98.06 $\pm$ 0.38 & 98.98 $\pm$ 0.28\\
BEiT-3 Large ft. COCO & 86.24 $\pm$ 0.95 & 97.26 $\pm$ 0.45 & 98.72 $\pm$ 0.31 & \textbf{86.64} $\pm$ \textbf{0.94} & \textbf{97.46} $\pm$ \textbf{0.44} & \textbf{98.92} $\pm$ \textbf{0.29} & 86.28 $\pm$ 0.95 & 97.24 $\pm$ 0.45 & 98.64 $\pm$ 0.32\\
\bottomrule
\end{tabular}%
}
\caption{\textbf{Full Flickr30k Image Retrieval Results for \nnd.} We report recall percentage with bootstrapped 95\% confidence intervals.} 
\label{tab:flickr30k_image_retrieval_results} 
\end{table*}

\begin{table*}[ht]
\centering
\resizebox{16cm}{!}{
\begin{tabular}{l|ccc|ccc|ccc} 

\toprule
& \multicolumn{3}{c|}{Flickr} & \multicolumn{3}{c|}{Flickr, \nnd\ w/ Flickr} & \multicolumn{3}{c}{Flickr, \nnd\ w/ COCO}\\ 
& R@1 & R@5 & R@10 & R@1 & R@5 & R@10 & R@1 & R@5 & R@10 \\ 
\midrule
CLIP & 79.30 $\pm$ 2.51 & 95.00 $\pm$ 1.35 & \textbf{98.10} $\pm$ \textbf{0.85} & \textbf{81.50} $\pm$ \textbf{2.41} & \textbf{95.70} $\pm$ \textbf{1.26} & 97.90 $\pm$ 0.89 & 79.70 $\pm$ 2.49 & 95.50 $\pm$ 1.28 & 98.00 $\pm$ 0.87\\
CLIP ft. Flickr & 85.70 $\pm$ 2.17 & \textbf{96.90} $\pm$ \textbf{1.07} & \textbf{98.70} $\pm$ \textbf{0.70} & \textbf{87.60} $\pm$ \textbf{2.04} & \textbf{96.90} $\pm$ \textbf{1.07} & 98.60 $\pm$ 0.73 & 87.30 $\pm$ 2.06 & \textbf{96.90} $\pm$ \textbf{1.07} & 98.60 $\pm$ 0.73\\
CLIP ft. COCO & 82.10 $\pm$ 2.38 & \textbf{95.90} $\pm$ \textbf{1.23} & 98.20 $\pm$ 0.82 & \textbf{83.00} $\pm$ \textbf{2.33} & 95.80 $\pm$ 1.24 & \textbf{98.50} $\pm$ \textbf{0.75} & 82.70 $\pm$ 2.34 & 95.80 $\pm$ 1.24 & 98.30 $\pm$ 0.80\\
BLIP & 93.70 $\pm$ 1.51 & 99.50 $\pm$ 0.44 & 99.90 $\pm$ 0.20 & \textbf{95.70} $\pm$ \textbf{1.26} & 99.50 $\pm$ 0.44 & 99.90 $\pm$ 0.20 & 94.50 $\pm$ 1.41 & \textbf{99.70} $\pm$ \textbf{0.34} & \textbf{100.00} $\pm$ \textbf{0.00}\\
BLIP ft. Flickr & 93.40 $\pm$ 1.54 & 99.50 $\pm$ 0.44 & 99.80 $\pm$ 0.28 & \textbf{95.40} $\pm$ \textbf{1.30} & 99.60 $\pm$ 0.39 & \textbf{99.90} $\pm$ \textbf{0.20} & 94.90 $\pm$ 1.36 & \textbf{99.80} $\pm$ \textbf{0.28} & \textbf{99.90} $\pm$ \textbf{0.20}\\
ALBEF ft. Flickr & 92.40 $\pm$ 1.64 & \textbf{99.10} $\pm$ \textbf{0.59} & 99.70 $\pm$ 0.34 & \textbf{92.70} $\pm$ \textbf{1.61} & 98.90 $\pm$ 0.65 & \textbf{99.80} $\pm$ \textbf{0.28} & 92.30 $\pm$ 1.65 & 99.00 $\pm$ 0.62 & \textbf{99.80} $\pm$ \textbf{0.28}\\
ALBEF ft. COCO & 87.30 $\pm$ 2.06 & 98.30 $\pm$ 0.80 & 99.20 $\pm$ 0.55 & \textbf{91.10} $\pm$ \textbf{1.76} & \textbf{99.30} $\pm$ \textbf{0.52} & \textbf{99.70} $\pm$ \textbf{0.34} & 89.60 $\pm$ 1.89 & 98.90 $\pm$ 0.65 & 99.60 $\pm$ 0.39\\
SigLIP & 89.00 $\pm$ 1.94 & 98.00 $\pm$ 0.87 & 99.30 $\pm$ 0.52 & \textbf{91.40} $\pm$ \textbf{1.74} & \textbf{98.60} $\pm$ \textbf{0.73} & \textbf{99.60} $\pm$ \textbf{0.39} & 90.30 $\pm$ 1.83 & 98.30 $\pm$ 0.80 & 99.20 $\pm$ 0.55\\
BEiT-3 & 89.10 $\pm$ 1.93 & 98.60 $\pm$ 0.73 & 99.20 $\pm$ 0.55 & \textbf{91.40} $\pm$ \textbf{1.74} & \textbf{98.90} $\pm$ \textbf{0.65} & 99.40 $\pm$ 0.48 & 90.60 $\pm$ 1.81 & 98.60 $\pm$ 0.73 & \textbf{99.50} $\pm$ \textbf{0.44}\\
BEiT-3 ft. Flickr & \textbf{96.30} $\pm$ \textbf{1.17} & \textbf{99.70} $\pm$ \textbf{0.34} & \textbf{100.00} $\pm$ \textbf{0.00} & 94.80 $\pm$ 1.38 & \textbf{99.70} $\pm$ \textbf{0.34} & \textbf{100.00} $\pm$ \textbf{0.00} & 94.70 $\pm$ 1.39 & 99.40 $\pm$ 0.48 & \textbf{100.00} $\pm$ \textbf{0.00}\\
BEiT-3 ft. COCO & 93.60 $\pm$ 1.52 & 99.30 $\pm$ 0.52 & 99.80 $\pm$ 0.28 & \textbf{95.40} $\pm$ \textbf{1.30} & \textbf{99.60} $\pm$ \textbf{0.39} & \textbf{99.90} $\pm$ \textbf{0.20} & 95.10 $\pm$ 1.34 & 99.30 $\pm$ 0.52 & \textbf{99.90} $\pm$ \textbf{0.20}\\
BEiT-3 Large & 91.10 $\pm$ 1.76 & 99.00 $\pm$ 0.62 & 99.60 $\pm$ 0.39 & \textbf{93.60} $\pm$ \textbf{1.52} & \textbf{99.30} $\pm$ \textbf{0.52} & \textbf{99.70} $\pm$ \textbf{0.34} & 92.50 $\pm$ 1.63 & 98.90 $\pm$ 0.65 & 99.60 $\pm$ 0.39\\
BEiT-3 Large ft. Flickr & 97.20 $\pm$ 1.02 & \textbf{100.00} $\pm$ \textbf{0.00} & \textbf{100.00} $\pm$ \textbf{0.00} & \textbf{97.30} $\pm$ \textbf{1.00} & \textbf{100.00} $\pm$ \textbf{0.00} & \textbf{100.00} $\pm$ \textbf{0.00} & 97.00 $\pm$ 1.06 & 99.90 $\pm$ 0.20 & \textbf{100.00} $\pm$ \textbf{0.00}\\
BEiT-3 Large ft. COCO & 95.50 $\pm$ 1.28 & 99.70 $\pm$ 0.34 & 99.80 $\pm$ 0.28 & \textbf{96.10} $\pm$ \textbf{1.20} & \textbf{99.90} $\pm$ \textbf{0.20} & \textbf{100.00} $\pm$ \textbf{0.00} & 95.90 $\pm$ 1.23 & 99.80 $\pm$ 0.28 & 99.90 $\pm$ 0.20\\
\bottomrule
\end{tabular}%
}
\caption{\textbf{Full Flickr30k Text Retrieval Results for \nnd.} We report recall percentage with bootstrapped 95\% confidence intervals.} 
\label{tab:flickr30k_text_retrieval_results} 
\end{table*}

\begin{table*}[ht]
\centering
\resizebox{16cm}{!}{
\begin{tabular}{l|ccc|ccc|ccc} 

\toprule
& \multicolumn{3}{c|}{COCO} & \multicolumn{3}{c|}{COCO, \nnd\ w/ Flickr} & \multicolumn{3}{c}{COCO, \nnd\ w/ COCO}\\ 
& R@1 & R@5 & R@10 & R@1 & R@5 & R@10 & R@1 & R@5 & R@10 \\ 
\midrule
CLIP & 30.45 $\pm$ 0.57 & 54.78 $\pm$ 0.62 & 66.23 $\pm$ 0.59 & 33.88 $\pm$ 0.59 & 59.12 $\pm$ 0.61 & 69.84 $\pm$ 0.57 & \textbf{37.76} $\pm$ \textbf{0.6} & \textbf{63.11} $\pm$ \textbf{0.6} & \textbf{73.46} $\pm$ \textbf{0.55}\\
BLIP & 62.72 $\pm$ 0.6 & 85.16 $\pm$ 0.44 & 91.32 $\pm$ 0.35 & 63.1 $\pm$ 0.6 & 85.28 $\pm$ 0.44 & 91.52 $\pm$ 0.35 & \textbf{64.49} $\pm$ \textbf{0.59} & \textbf{86.33} $\pm$ \textbf{0.43} & \textbf{92.02} $\pm$ \textbf{0.34}\\
CLIP ft F & 35.58 $\pm$ 0.59 & 61.27 $\pm$ 0.6 & 71.69 $\pm$ 0.56 & 36.62 $\pm$ 0.6 & 62.17 $\pm$ 0.6 & 72.34 $\pm$ 0.55 & \textbf{40.36} $\pm$ \textbf{0.61} & \textbf{65.9} $\pm$ \textbf{0.59} & \textbf{76.14} $\pm$ \textbf{0.53}\\
BLIP ft F & 56.47 $\pm$ 0.61 & 81.18 $\pm$ 0.48 & 88.45 $\pm$ 0.4 & 57.65 $\pm$ 0.61 & 81.4 $\pm$ 0.48 & 88.62 $\pm$ 0.39 & \textbf{60.03} $\pm$ \textbf{0.61} & \textbf{83.11} $\pm$ \textbf{0.46} & \textbf{89.66} $\pm$ \textbf{0.38}\\
ALBEF ft F & 52.56 $\pm$ 0.62 & 79.07 $\pm$ 0.5 & 87.05 $\pm$ 0.42 & 53.56 $\pm$ 0.62 & 79.32 $\pm$ 0.5 & 87.3 $\pm$ 0.41 & \textbf{56.89} $\pm$ \textbf{0.61} & \textbf{82.14} $\pm$ \textbf{0.47} & \textbf{89.04} $\pm$ \textbf{0.39}\\
ALBEF ft C & 59.76 $\pm$ 0.61 & 84.28 $\pm$ 0.45 & 90.56 $\pm$ 0.36 & 60.24 $\pm$ 0.61 & 84.54 $\pm$ 0.45 & 91.0 $\pm$ 0.35 & \textbf{62.92} $\pm$ \textbf{0.6} & \textbf{85.97} $\pm$ \textbf{0.43} & \textbf{91.74} $\pm$ \textbf{0.34}\\
CLIP ft C & 45.92 $\pm$ 0.62 & 73.2 $\pm$ 0.55 & 82.56 $\pm$ 0.47 & 46.28 $\pm$ 0.62 & 73.02 $\pm$ 0.55 & 82.55 $\pm$ 0.47 & \textbf{47.93} $\pm$ \textbf{0.62} & \textbf{74.17} $\pm$ \textbf{0.54} & \textbf{82.86} $\pm$ \textbf{0.47}\\
SigLIP & 47.18 $\pm$ 0.62 & 72.08 $\pm$ 0.56 & 80.58 $\pm$ 0.49 & 48.72 $\pm$ 0.62 & 73.2 $\pm$ 0.55 & 81.78 $\pm$ 0.48 & \textbf{50.72} $\pm$ \textbf{0.62} & \textbf{74.99} $\pm$ \textbf{0.54} & \textbf{82.7} $\pm$ \textbf{0.47}\\
BEiT-3 base & 47.64 $\pm$ 0.62 & 72.54 $\pm$ 0.55 & 81.2 $\pm$ 0.48 & 48.22 $\pm$ 0.62 & 73.31 $\pm$ 0.55 & 81.86 $\pm$ 0.48 & \textbf{50.83} $\pm$ \textbf{0.62} & \textbf{75.56} $\pm$ \textbf{0.53} & \textbf{83.42} $\pm$ \textbf{0.46}\\
BEiT-3 ft on F & 53.59 $\pm$ 0.62 & 77.98 $\pm$ 0.51 & 85.71 $\pm$ 0.43 & 53.99 $\pm$ 0.62 & 78.31 $\pm$ 0.51 & 85.96 $\pm$ 0.43 & \textbf{56.24} $\pm$ \textbf{0.61} & \textbf{80.07} $\pm$ \textbf{0.5} & \textbf{87.25} $\pm$ \textbf{0.41}\\
BEiT-3 ft on C & 61.91 $\pm$ 0.6 & 85.15 $\pm$ 0.44 & 91.49 $\pm$ 0.35 & 61.8 $\pm$ 0.6 & 84.97 $\pm$ 0.44 & 91.28 $\pm$ 0.35 & \textbf{62.3} $\pm$ \textbf{0.6} & \textbf{85.22} $\pm$ \textbf{0.44} & \textbf{91.58} $\pm$ \textbf{0.34}\\
BEiT-3 large & 49.36 $\pm$ 0.62 & 73.64 $\pm$ 0.55 & 81.85 $\pm$ 0.48 & 50.18 $\pm$ 0.62 & 74.27 $\pm$ 0.54 & 82.42 $\pm$ 0.47 & \textbf{52.54} $\pm$ \textbf{0.62} & \textbf{76.44} $\pm$ \textbf{0.53} & \textbf{84.13} $\pm$ \textbf{0.45}\\
BEiT-3 large ft on F & 56.43 $\pm$ 0.61 & 80.4 $\pm$ 0.49 & 87.72 $\pm$ 0.41 & 56.9 $\pm$ 0.61 & 80.54 $\pm$ 0.49 & 87.72 $\pm$ 0.41 & \textbf{58.97} $\pm$ \textbf{0.61} & \textbf{81.69} $\pm$ \textbf{0.48} & \textbf{88.71} $\pm$ \textbf{0.39}\\
BEiT-3 large ft on C & 63.85 $\pm$ 0.6 & 86.41 $\pm$ 0.42 & 92.31 $\pm$ 0.33 & 63.76 $\pm$ 0.6 & 86.18 $\pm$ 0.43 & 92.18 $\pm$ 0.33 & \textbf{64.54} $\pm$ \textbf{0.59} & \textbf{86.42} $\pm$ \textbf{0.42} & \textbf{92.32} $\pm$ \textbf{0.33}\\
\bottomrule
\end{tabular}%
}
\caption{\textbf{Full COCO Image Retrieval Results for \nnd.} We report recall percentage with bootstrapped 95\% confidence intervals.} 
\label{tab:coco_image_retrieval_results} 
\end{table*}

\begin{table*}[ht]
\centering
\resizebox{16cm}{!}{
\begin{tabular}{l|ccc|ccc|ccc} 

\toprule
& \multicolumn{3}{c|}{COCO} & \multicolumn{3}{c|}{COCO, \nnd\ w/ Flickr} & \multicolumn{3}{c}{COCO, \nnd\ w/ COCO}\\ 
& R@1 & R@5 & R@10 & R@1 & R@5 & R@10 & R@1 & R@5 & R@10 \\ 
\midrule
CLIP & 50.02 $\pm$ 1.39 & 74.84 $\pm$ 1.20 & 83.18 $\pm$ 1.04 & 51.74 $\pm$ 1.39 & 75.94 $\pm$ 1.18 & 83.86 $\pm$ 1.02 & \textbf{54.16} $\pm$ \textbf{1.38} & \textbf{77.60} $\pm$ \textbf{1.16} & \textbf{85.46} $\pm$ \textbf{0.98}\\
CLIP ft. Flickr & 53.74 $\pm$ 1.38 & 76.36 $\pm$ 1.18 & 84.36 $\pm$ 1.01 & 53.68 $\pm$ 1.38 & 76.48 $\pm$ 1.18 & 84.80 $\pm$ 1.00 & \textbf{56.86} $\pm$ \textbf{1.37} & \textbf{79.14} $\pm$ \textbf{1.13} & \textbf{86.68} $\pm$ \textbf{0.94}\\
CLIP ft. COCO & 63.74 $\pm$ 1.33 & 85.84 $\pm$ 0.97 & 91.54 $\pm$ 0.77 & 64.06 $\pm$ 1.33 & 85.74 $\pm$ 0.97 & 91.54 $\pm$ 0.77 & \textbf{65.44} $\pm$ \textbf{1.32} & \textbf{86.20} $\pm$ \textbf{0.96} & \textbf{91.92} $\pm$ \textbf{0.76}\\
BLIP & 79.62 $\pm$ 1.12 & 94.48 $\pm$ 0.63 & 97.20 $\pm$ 0.46 & 79.98 $\pm$ 1.11 & 94.70 $\pm$ 0.62 & 97.34 $\pm$ 0.45 & \textbf{82.68} $\pm$ \textbf{1.05} & \textbf{95.32} $\pm$ \textbf{0.59} & \textbf{97.86} $\pm$ \textbf{0.40}\\
BLIP ft. Flickr & 72.26 $\pm$ 1.24 & 90.34 $\pm$ 0.82 & 94.80 $\pm$ 0.62 & 74.88 $\pm$ 1.20 & 91.84 $\pm$ 0.76 & 95.88 $\pm$ 0.55 & \textbf{78.64} $\pm$ \textbf{1.14} & \textbf{93.28} $\pm$ \textbf{0.69} & \textbf{96.54} $\pm$ \textbf{0.51}\\
ALBEF ft. Flickr & 69.82 $\pm$ 1.27 & 91.16 $\pm$ 0.79 & 95.32 $\pm$ 0.59 & 71.10 $\pm$ 1.26 & 91.58 $\pm$ 0.77 & 95.88 $\pm$ 0.55 & \textbf{74.82} $\pm$ \textbf{1.20} & \textbf{92.60} $\pm$ \textbf{0.73} & \textbf{96.24} $\pm$ \textbf{0.53}\\
ALBEF ft. COCO & 78.60 $\pm$ 1.14 & 94.82 $\pm$ 0.61 & 97.54 $\pm$ 0.43 & 79.06 $\pm$ 1.13 & 95.32 $\pm$ 0.59 & \textbf{97.78} $\pm$ \textbf{0.41} & \textbf{80.86} $\pm$ \textbf{1.09} & \textbf{95.50} $\pm$ \textbf{0.57} & 97.62 $\pm$ 0.42\\
SigLIP & 65.32 $\pm$ 1.32 & 86.22 $\pm$ 0.96 & 91.60 $\pm$ 0.77 & 67.04 $\pm$ 1.30 & 87.18 $\pm$ 0.93 & 92.48 $\pm$ 0.73 & \textbf{70.24} $\pm$ \textbf{1.27} & \textbf{88.12} $\pm$ \textbf{0.90} & \textbf{93.34} $\pm$ \textbf{0.69}\\
BEiT-3 & 61.12 $\pm$ 1.35 & 83.96 $\pm$ 1.02 & 90.86 $\pm$ 0.80 & 66.02 $\pm$ 1.31 & 87.06 $\pm$ 0.93 & 92.64 $\pm$ 0.72 & \textbf{69.26} $\pm$ \textbf{1.28} & \textbf{88.70} $\pm$ \textbf{0.88} & \textbf{93.24} $\pm$ \textbf{0.70}\\
BEiT-3 ft. Flickr & 72.02 $\pm$ 1.24 & 90.50 $\pm$ 0.81 & 94.72 $\pm$ 0.62 & 72.64 $\pm$ 1.24 & 90.84 $\pm$ 0.80 & 94.90 $\pm$ 0.61 & \textbf{75.12} $\pm$ \textbf{1.20} & \textbf{92.20} $\pm$ \textbf{0.74} & \textbf{95.68} $\pm$ \textbf{0.56}\\
BEiT-3 ft. COCO & 80.72 $\pm$ 1.09 & \textbf{95.60} $\pm$ \textbf{0.57} & \textbf{98.12} $\pm$ \textbf{0.38} & 80.58 $\pm$ 1.10 & 95.58 $\pm$ 0.57 & 97.94 $\pm$ 0.39 & \textbf{80.82} $\pm$ \textbf{1.09} & 95.50 $\pm$ 0.57 & 97.96 $\pm$ 0.39\\
BEiT-3 Large & 63.26 $\pm$ 1.34 & 85.60 $\pm$ 0.97 & 91.70 $\pm$ 0.76 & 67.84 $\pm$ 1.29 & 88.02 $\pm$ 0.90 & 92.98 $\pm$ 0.71 & \textbf{70.74} $\pm$ \textbf{1.26} & \textbf{89.30} $\pm$ \textbf{0.86} & \textbf{94.32} $\pm$ \textbf{0.64}\\
BEiT-3 Large ft. Flickr & 74.32 $\pm$ 1.21 & 92.06 $\pm$ 0.75 & 95.82 $\pm$ 0.55 & 74.64 $\pm$ 1.21 & 91.94 $\pm$ 0.75 & 95.84 $\pm$ 0.55 & \textbf{78.72} $\pm$ \textbf{1.13} & \textbf{93.30} $\pm$ \textbf{0.69} & \textbf{96.62} $\pm$ \textbf{0.50}\\
BEiT-3 Large ft. COCO & 82.10 $\pm$ 1.06 & \textbf{96.12} $\pm$ \textbf{0.54} & 98.40 $\pm$ 0.35 & 82.16 $\pm$ 1.06 & 95.96 $\pm$ 0.55 & \textbf{98.58} $\pm$ \textbf{0.33} & \textbf{83.00} $\pm$ \textbf{1.04} & 96.04 $\pm$ 0.54 & 98.40 $\pm$ 0.35\\
\bottomrule
\end{tabular}%
}
\caption{\textbf{Full COCO Text Retrieval Results for \nnd.} We report recall percentage with bootstrapped 95\% confidence intervals.} 
\label{tab:coco_text_retrieval_results} 
\end{table*}

\section{Ablation Study}
\label{sec:appendix_ablation}

In some scenarios, it is possible that one may not have access to a very large reference query dataset. To simulate the performance of \nnd\ and other baselines under this constraint, in Table~\ref{tab:flickr30k_ablation} and \ref{tab:coco_ablation}, we show the retrieval scores when only a subset of the Flickr30k/COCO queries are used as the reference dataset. We find that \nnd\ substantially improves beyond the base model even for ablated datasets.

\begin{table}[H]
\centering
\resizebox{\columnwidth}{!}{
\begin{tabular}{lccccc}
\toprule
Model & Original & NNN (full) & NNN (50\%) & NNN (20\%) & NNN (10\%) \\
\midrule
CLIP & 58.82 & 64.94 & 64.80 & 64.60 & 64.84\\
CLIP ft. Flickr & 72.80 & 74.06 & 73.86 & 74.14 & 74.42\\
CLIP ft. COCO & 67.40 & 69.64 & 69.18 & 68.86 & 68.86\\
BLIP ft. Flickr & 83.58 & 84.48 & 84.44 & 84.32 & 84.18\\
BLIP ft. COCO & 82.12 & 83.32 & 83.28 & 82.80 & 83.04\\
ALBEF ft. Flickr & 79.50 & 81.02 & 80.84 & 80.26 & 80.10\\
ALBEF ft. COCO & 74.54 & 76.86 & 77.04 & 76.60 & 76.48\\
SigLIP & 74.62 & 76.82 & 76.70 & 76.54 & 76.40\\
BEiT-3 & 75.52 & 76.88 & 76.92 & 76.66 & 76.70\\
BEiT-3 ft. Flickr & 86.12 & 86.36 & 86.10 & 86.00 & 86.06\\
BEiT-3 ft. COCO & 82.90 & 83.72 & 83.46 & 83.48 & 83.16\\
BEiT-3 Large & 77.80 & 78.94 & 78.68 & 78.54 & 78.44\\
BEiT-3 Large ft. Flickr & 88.04 & 87.96 & 87.90 & 87.82 & 87.88\\
BEiT-3 Large ft. COCO & 86.24 & 86.98 & 86.66 & 86.64 & 86.56\\
\bottomrule
\end{tabular}
}
\vspace{1mm}
\caption{\textbf{Flickr30k ablation studies (Image Retrieval@1).}} 
\label{tab:flickr30k_ablation} 
\end{table}

\begin{table}[H]
\centering
\resizebox{\columnwidth}{!}{
\begin{tabular}{lccccc}
\toprule
Model & Original & NNN (full) & NNN (50\%) & NNN (20\%) & NNN (10\%) \\
\midrule
CLIP & 79.30 & 81.90 & 81.90 & 81.20 & 81.60\\
CLIP ft. Flickr & 85.70 & 87.30 & 87.00 & 87.30 & 87.10\\
CLIP ft. COCO & 82.10 & 82.10 & 82.20 & 82.80 & 82.50\\
BLIP ft. Flickr & 93.40 & 95.00 & 95.40 & 95.20 & 95.50\\
BLIP ft. COCO & 93.70 & 95.20 & 95.20 & 95.30 & 95.30\\
ALBEF ft. Flickr & 92.40 & 92.80 & 92.80 & 92.60 & 92.60\\
ALBEF ft. COCO & 87.30 & 90.50 & 90.30 & 90.00 & 89.50\\
SigLIP & 89.00 & 91.20 & 91.20 & 91.30 & 91.10\\
BEiT-3 & 89.10 & 91.50 & 91.70 & 91.80 & 90.90\\
BEiT-3 ft. Flickr & 96.30 & 95.40 & 96.00 & 95.60 & 95.80\\
BEiT-3 ft. COCO & 93.60 & 95.40 & 94.90 & 95.30 & 94.60\\
BEiT-3 Large & 91.10 & 93.60 & 93.30 & 93.20 & 91.60\\
BEiT-3 Large ft. Flickr & 97.20 & 97.40 & 97.20 & 97.20 & 97.10\\
BEiT-3 Large ft. COCO & 95.50 & 95.20 & 95.40 & 95.30 & 95.50\\
\bottomrule
\end{tabular}
}
\vspace{1mm}
\caption{\textbf{Flickr30k ablation studies (Text Retrieval@1).}} 
\label{tab:flickr30k_ablation} 
\end{table}

\begin{table}[H]
\centering
\resizebox{\columnwidth}{!}{
\begin{tabular}{lccccc}
\toprule
Model & Original & NNN (full) & NNN (50\%) & NNN (20\%) & NNN (10\%) \\
\midrule
CLIP & 30.43 & 37.74 & 37.48 & 37.53 & 37.43\\
CLIP ft. Flickr & 35.56 & 40.13 & 40.17 & 40.12 & 40.28\\
CLIP ft. COCO & 45.89 & 47.90 & 47.70 & 47.39 & 47.35\\
BLIP ft. Flickr & 56.44 & 60.12 & 60.00 & 59.70 & 59.56\\
BLIP ft. COCO & 62.68 & 64.45 & 64.35 & 64.44 & 64.14\\
ALBEF ft. Flickr & 52.53 & 57.09 & 56.88 & 56.67 & 56.40\\
ALBEF ft. COCO & 59.73 & 62.88 & 62.82 & 62.66 & 62.43\\
SigLIP & 47.15 & 50.70 & 50.72 & 50.24 & 50.15\\
BEiT-3 & 47.62 & 50.81 & 50.80 & 50.64 & 50.50\\
BEiT-3 ft. Flickr & 53.57 & 56.19 & 56.16 & 55.91 & 55.97\\
BEiT-3 ft. COCO & 61.88 & 62.54 & 62.46 & 62.34 & 62.26\\
BEiT-3 Large & 49.34 & 52.52 & 52.42 & 52.25 & 52.09\\
BEiT-3 Large ft. Flickr & 56.41 & 58.91 & 58.88 & 58.88 & 58.66\\
BEiT-3 Large ft. COCO & 63.83 & 64.14 & 64.13 & 64.20 & 64.07\\
\bottomrule
\end{tabular}
}
\vspace{1mm}
\caption{\textbf{COCO ablation studies (Image Retrieval@1).}} 
\label{tab:coco_ablation} 
\end{table}

\begin{table}[H]
\centering
\resizebox{\columnwidth}{!}{
\begin{tabular}{lccccc}
\toprule
Model & Original & NNN (full) & NNN (50\%) & NNN (20\%) & NNN (10\%) \\
\midrule
CLIP & 50.02 & 53.94 & 53.88 & 53.66 & 53.66\\
CLIP ft. Flickr & 53.74 & 56.86 & 56.70 & 56.44 & 56.24\\
CLIP ft. COCO & 63.74 & 65.44 & 65.40 & 65.26 & 64.44\\
BLIP ft. Flickr & 72.26 & 78.64 & 78.04 & 78.30 & 78.24\\
BLIP ft. COCO & 79.62 & 82.70 & 82.42 & 82.46 & 82.10\\
ALBEF ft. Flickr & 69.82 & 75.16 & 74.64 & 74.44 & 74.66\\
ALBEF ft. COCO & 78.60 & 81.22 & 81.00 & 80.68 & 80.26\\
SigLIP & 65.32 & 70.24 & 70.42 & 69.86 & 69.98\\
BEiT-3 & 61.12 & 69.26 & 69.30 & 69.12 & 69.00\\
BEiT-3 ft. Flickr & 72.02 & 75.50 & 75.16 & 75.22 & 75.14\\
BEiT-3 ft. COCO & 80.72 & 81.58 & 81.30 & 81.26 & 81.26\\
BEiT-3 Large & 63.26 & 70.74 & 70.84 & 71.08 & 70.72\\
BEiT-3 Large ft. Flickr & 74.32 & 78.64 & 78.42 & 77.92 & 77.34\\
BEiT-3 Large ft. COCO & 82.10 & 82.92 & 82.86 & 82.72 & 82.72\\
\bottomrule
\end{tabular}
}
\vspace{1mm}
\caption{\textbf{COCO ablation studies (Text Retrieval@1).}} 
\label{tab:coco_ablation} 
\end{table}


\newpage

\begin{figure}[H]
    \centering
    \includegraphics[width=\linewidth]{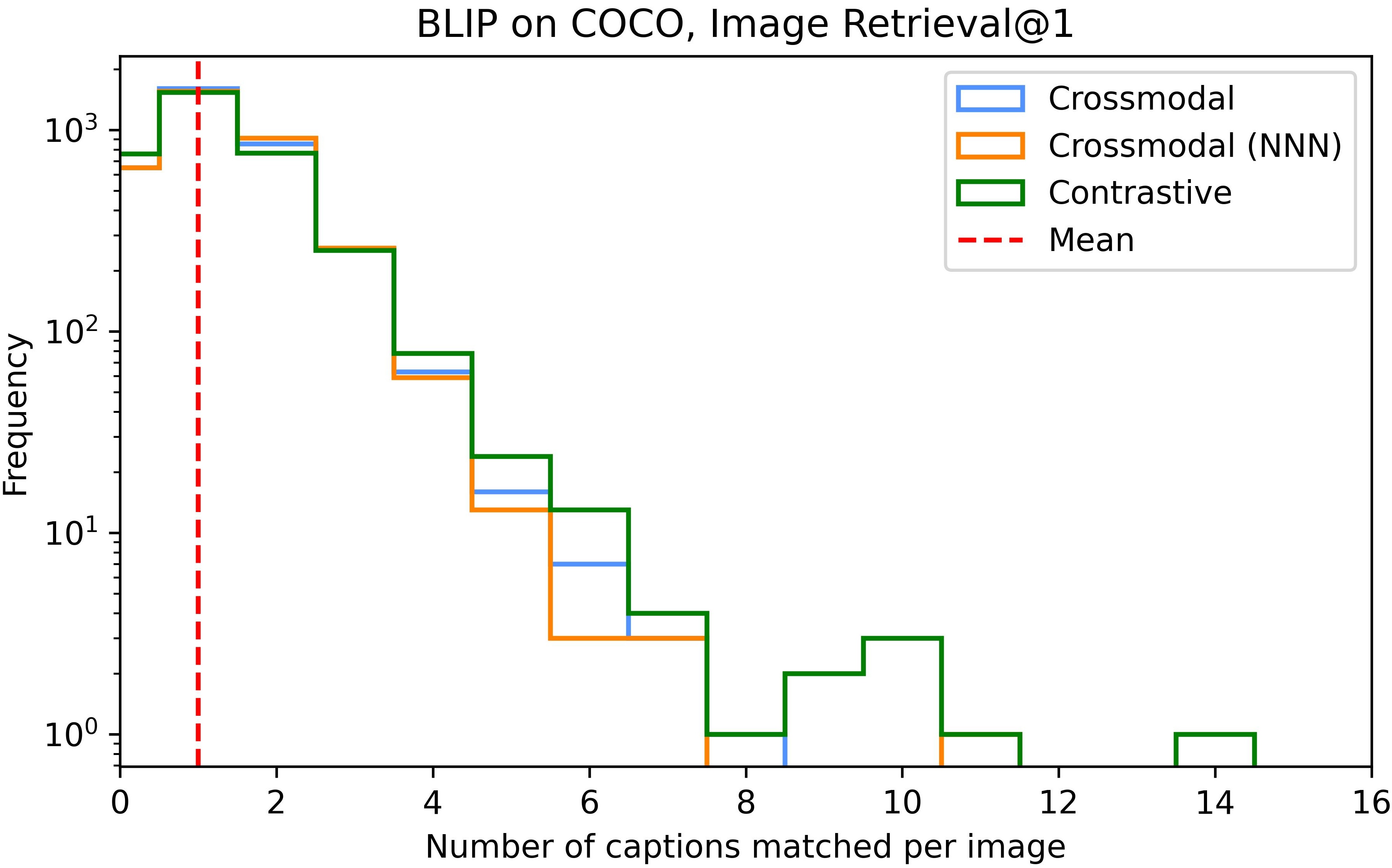}
    \vspace{-6mm}
    \caption{\textbf{Distribution of COCO captions matched to each image during image retrieval for BLIP crossmodal} Applying \nnd\  to the cross-attention model does not significantly affect the distribution: a Kolmogorov-Smirnov test has a p-value of 0.846. (One caption was chosen per image due to compute constraints.)
    }
    \label{fig:blip_itm}
\end{figure}

\section{Crossmodal attention}
\label{sec:appendix_crossmodal}
We find that \nnd\ consistently increases retrieval accuracy in contrastive models, but does not significantly improve cross-attention models: for the image-text matching version of BLIP on COCO, Image Recall@1 improves from $66.16\%$ to $66.24\%$  (Figure~\ref{fig:blip_itm}).

\section{Image and caption bias (extended results)}
\label{sec:appendix_imagecaptionbias}

\begin{figure*}[t!]
    \centering
    \includegraphics[width=\textwidth]{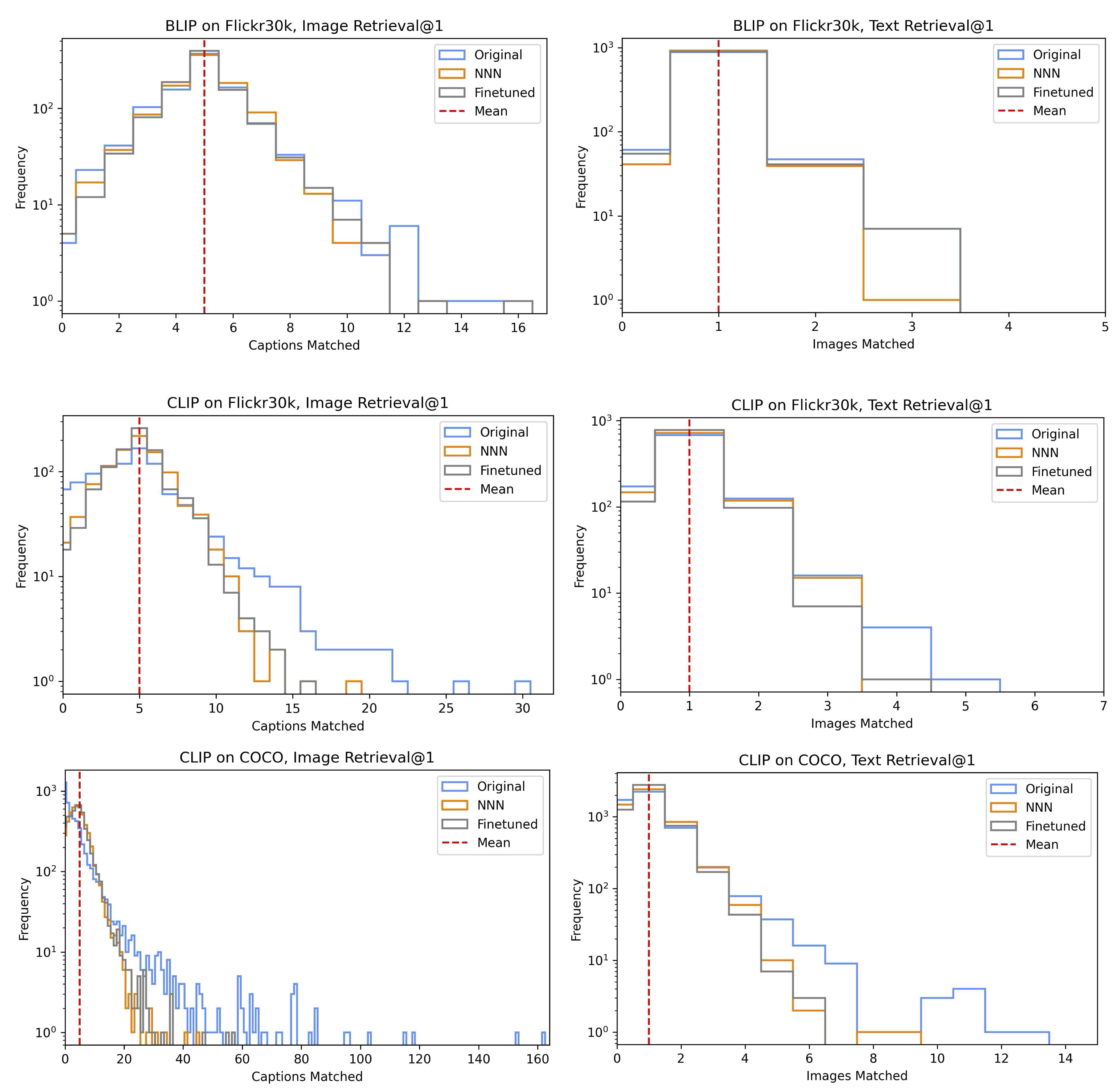}
    \vspace{-6mm}
    \caption{\textbf{Distribution of captions matched per image for image retrieval (left), and images matched per caption for text retrieval (right).}}
    \vspace{10mm}
    \label{fig:more_histograms}
\end{figure*}

In Figure~\ref{fig:more_histograms}, we show more examples of reducing hubness using \nnd\ for both text retrieval and image retrieval. The effect is more observable in image retrieval as there are 5 times more captions than images. 

